\pdfoutput=1

\documentclass[11pt]{article}
\usepackage[final]{ACL2023}

\usepackage[utf8]{inputenc} %
\usepackage{hyperref}       %
\usepackage{url}            %
\usepackage{nicefrac}       %

\newcommand{\isoCost}{isoCost\xspace}

\usepackage{times}
\usepackage{latexsym}

\usepackage{tcolorbox}
\tcbuselibrary{breakable}

\usepackage{microtype}

\usepackage{amsmath}
\usepackage{amsthm}
\usepackage{amsfonts}
\usepackage{amssymb}
\usepackage{mathtools}

\mathchardef\mhyphen="2D

\DeclarePairedDelimiterX{\infdivx}[2]{(}{)}{%
  #1\;\delimsize\|\;#2%
}

\usepackage{todonotes}
\makeatletter
\newcommand*\iftodonotes{\if@todonotes@disabled\expandafter\@secondoftwo\else\expandafter\@firstoftwo\fi}
\makeatother

\definecolor{lightblue}{rgb}{0.7,0.85,1}

\usepackage{tikz}
\usetikzlibrary{bayesnet}
\usepackage{graphicx}
\usepackage{booktabs}
\usepackage{tabularx}
\usepackage{tabulary}
\usepackage{colortbl}
\usepackage{xspace}
\usepackage{xcolor}
\newcolumntype{Y}{>{\centering\arraybackslash}X}
\usepackage{multirow}

\usepackage[capitalize,noabbrev]{cleveref}

\usepackage{soul}

\usepackage{subcaption}

\title{The Sparse Frontier: Sparse Attention Trade-offs in Transformer LLMs}

\author{Piotr Nawrot\footnotemark[1]
\\ University of Edinburgh \\\And Robert Li \\ Cohere \\\And Renjie Huang \\ Cohere \AND Sebastian Ruder\footnotemark[2] \\ Meta \\\And Kelly Marchisio \\ Cohere \\\And Edoardo M. Ponti \\ University of Edinburgh}

\begin{document}
\maketitle
\renewcommand*{\thefootnote}{\fnsymbol{footnote}}
\footnotetext[1]{Research conducted during an internship at Cohere. Correspondence email: piotr.nawrot@ed.ac.uk}
\footnotetext[2]{Work done prior to joining Meta.}

\begin{abstract}
Sparse attention offers a promising strategy to extend long-context capabilities in Transformer LLMs, yet its efficiency--accuracy trade-offs remain unclear due to the lack of comprehensive evaluation. We address this gap with the largest-scale empirical analysis to date of training-free sparse attention, evaluating six methods across multiple model families and sizes, sequences up to 128K tokens, and sparsity levels up to 0.95 (i.e., $1/20$ attention budget) on nine diverse tasks. We first organise the rapidly evolving landscape of sparse attention methods into a taxonomy along four design axes. Our analysis then yields actionable insights: 1) sparse attention is effective: larger sparse models outperform smaller dense ones at equivalent cost, improving the Pareto frontier; 2) for the training-free methods we study, fine-grained per-query importance estimation during prefilling remains impractical---due to both the cost of estimation and the lack of sparse kernels that translate fine-grained sparsity into wall-clock gains---forcing a task-dependent choice between global-to-token and block-to-block selection. Instead, during decoding, token-to-page selection becomes feasible, enabling better generalisation and higher sparsity tolerance; 3) longer sequences tolerate higher sparsity, suggesting that fixed-budget methods in production are suboptimal. Together, these findings provide practical guidance for deploying sparse attention and methodological recommendations for future evaluations. Our code is available at \url{https://github.com/PiotrNawrot/sparse-frontier}.

\end{abstract}

\section{Introduction}
\label{sec:introduction}

\begin{figure*}[!t]
\centering
\includegraphics[width=\textwidth]{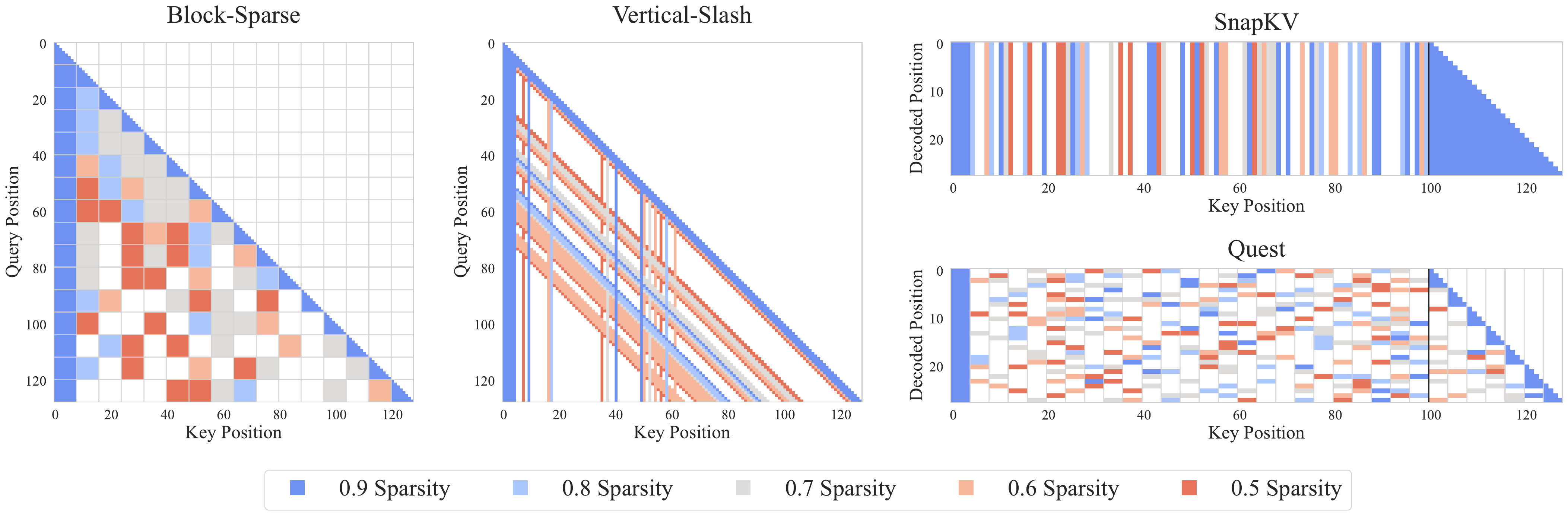}
\caption{Visualization of sparse attention patterns. Block-Sparse and Vertical-Slash operate during prefilling (showing query-key attention matrix), while SnapKV and Quest operate during decoding (showing decoded tokens attending to KV cache positions). Colors indicate different sparsity levels from 0.5 (red) to 0.9 (blue). The black vertical lines in SnapKV and Quest mark the prefill/decode boundary.}
\label{fig:sparse_attention_patterns}
\end{figure*}

The ability to model long sequences in large language models (LLMs) lies at the heart of long-context processing \citep{liu2025comprehensive} and inference-time scaling \citep{snell2024scaling,muennighoff2025s1}. The fundamental bottleneck for this ability is the self-attention mechanism \citep{Bahdanau2014NeuralMT, vaswani2017attention}: during the prefilling stage, its computational complexity scales quadratically with sequence length---hence, ballooning time-to-first-token and deployment cost \citep{Jiang2024MInference1A}. In the decoding phase, the key--value (KV) cache grows linearly with sequence length, and the need to load from memory this expanding cache for each generation step dominates the runtime \citep{Nawrot2024DynamicMC}.

Sparse attention mechanisms aim to address these challenges by approximating dense attention outputs with only a subset of query--key interactions \citep{Fu2024ChallengesID}. These span both training-based variants---such as DMS \citep{lancucki2025hyperscaling}, DeepSeek's NSA \citep{Yuan2025NativeSA}, and SWA used in OpenAI's gpt-oss and Google's Gemma 3---and training-free methods that operate directly on pretrained models, such as Vertical-Slash \citep{Jiang2024MInference1A} deployed in Qwen 2.5-1M \citep{yang2025qwen251m} and integrated into vLLM. Sparse attention has not only seen widespread adoption in industry, %
but also in the research community: over 150 papers with ``sparse attention'' in the title were submitted to arXiv between January 2025 and January 2026. %
Despite this popularity, the viability and robustness of sparse attention remain unclear due to a lack of comprehensive large-scale evaluation.

In this work, we carry out the largest-scale empirical analysis to date of training-free sparse attention methods, covering three model families (Qwen 2.5, Llama 3.1, Gemma 3) with sizes between 4B and 72B parameters, sequences between 16K to 128K tokens, and sparsity levels up to 0.95 (i.e., $1/20$ attention budget). To enable a controlled analysis, we first survey existing approaches, addressing the challenge of comparing rapidly evolving methods whose implementation details often obscure their core design principles. We distil these approaches into four key axes: units of sparsification (blocks/pages or verticals and slashes), importance estimation (fixed or context-aware), budget allocation across layers (uniform or adaptive), and KV cache management (eviction or full cache). Based on this taxonomy, we select six representative methods spanning these design dimensions and harmonise their implementations, allowing us to rigorously evaluate their distinct effects.

We focus specifically on training-free sparse attention because training-based alternatives require prohibitive computational resources and possibly access to proprietary training data \citep{Nawrot2024DynamicMC}. While this is a limitation, we expect insights from our training-free analysis to transfer to training-based methods given their similarity.\footnote{For instance, Quest \citep{Tang2024QuestQS} and NSA \citep{Yuan2025NativeSA} both use page-based selection for sparse decoding.} While previously \citet{Li2024SCBenchAK}, \citet{liu2025can}, and \citet{Yuan2024KVCC} provided a preliminary exploration of training-free methods, they covered limited configurations, specific use cases, or did not control for sequence lengths, hindering a systematic analysis (see \cref{app:prior_work} for a detailed comparison).

For our evaluation, we curate a benchmark suite of 9 long-context tasks designed to systematically probe the influence of key factors on sparse attention performance. These factors include diverse \textit{task types} (ranging from retrieval to multi-hop variable tracking and information aggregation), varying \textit{naturalness} of sequences (synthetic or natural language), and precisely controlled \textit{sequence lengths}. The importance of these dimensions is underscored by prior work indicating their significant impact on sparse attention effectiveness \citep{Chen2024MagicPIGLS, Liu2024ReAttentionTI}. Alongside established benchmarks \citep{Rajpurkar2018KnowWY,Pang2021QuALITYQA,tseng2016towards}, we introduce novel, more challenging tasks based on natural language story templates. These complement synthetic benchmarks like RULER \citep{Hsieh2024RULERWT}, whose results may fail to extrapolate to realistic data, by evaluating core skills in a controllable yet realistic natural language setting. All this provides us with a toolbox to address fundamental questions that currently remain unresolved:

\textit{RQ1: Is sparse attention effective?} (\cref{sec:results_isoflop}) An \isoCost analysis reveals that sparsification enables larger sparse models to outperform smaller dense ones at equivalent cost (i.e., FLOPs during prefilling and memory reads during decoding). For long sequences, only high-sparsity configurations lie on the Pareto frontier.

\textit{RQ2: Which sparse attention method should practitioners use?} (\cref{sec:results_individual}) Prefill and decoding phases display different trends. Among the training-free methods we evaluate, prefilling forces a choice between fine-grained global selection (Vertical-Slash) and block-to-block selection (Block-Sparse)---neither of which generalises across all tasks---because no existing training-free pipeline pairs an efficient fine-grained importance estimator with a kernel that exploits the resulting sparsity. During decoding, per-query selection is cheap and token-to-page methods (e.g., Quest) enable greater flexibility and higher compression tolerance.

\textit{RQ3: How does the sequence length affect tolerance to sparse attention?} (\cref{sec:results_seq_length}) Longer sequences permit higher sparsity while maintaining accuracy, consistently across model families. This suggests that fixed-budget methods deployed in production are suboptimal, and future designs should adapt sparsity to sequence length.

Overall, our findings provide practical guidance for deploying sparse attention and methodological recommendations for future evaluations in this rapidly evolving field.

\section{Training-Free Sparse Attention}
\label{sec:background}

The self-attention mechanism computes query $Q$, key $K$, and value $V$ representations from an input sequence $X \in \mathbb{R}^{n \times d}$. The output $O_i$ for the $i$-th token is a weighted sum of values, $O_i = \sum_{j=1}^{n} A_{ij}V_j$, where the attention weights $A_{ij}$ are derived from scaled dot-products between queries and keys: $A_i = \text{softmax}(Q_i K^\top / \sqrt{d})$. We omit multi-head details for brevity.

Transformer-based text generation involves two phases. \textbf{Prefilling} processes the entire input sequence, computing the lower-triangular part of the $n \times n$ attention matrix $A$, leading to $O(n^2)$ complexity. \textbf{Decoding} generates tokens auto-regressively. While attention is $O(n)$ per step (single query), loading the expanding Key-Value (KV) cache from memory becomes the main bottleneck.

Sparse attention methods reduce these costs by computing only a subset of QK interactions, making $A$ sparse. This lowers computational load during prefilling and memory transfers during decoding. We quantify the effectiveness of these methods using \textbf{sparsity}---the fraction of non-computed QK interactions. Equivalently, sparsity of $1 - 1/k$ corresponds to retaining only a $1/k$ fraction of the attention interactions. For instance, sparsity 0.9 (or equivalently, $1/10$ attention budget) means computing only 10\% of the original QK interactions.

The speedup from sparse attention depends on how much of total cost is attention. Since attention scales quadratically while other components (MLP, embeddings) scale linearly with sequence length, attention dominates at longer contexts---yielding greater benefits from sparsification. Models with built-in architectural sparsity, such as sliding-window or linear attention layers, have lower baseline attention ratios and require longer sequences for additional sparsification to provide comparable gains (\cref{sec:cost_breakdown}).

We categorise training-free sparse attention methods along four axes: {unit of sparsification}, {importance estimation}, {budget allocation}, and {KV cache management}. We exclude token merging methods \citep{Wang2024ModelTY,Nawrot2024DynamicMC}, which do not rely on sparsity.

\subsection{Unit of Sparsification}

Sparse attention methods differ primarily in the structural units of the attention matrix they prune or retain. Common units include \textit{local windows} (contiguous regions around each query), \textit{vertical columns} (tokens globally available to all queries), \textit{slashes} (tokens at fixed offsets from each query), and \textit{blocks} (fixed-size tiles of the attention matrix, such as 64$\times$64 tokens). Larger structured units such as blocks or windows offer improved computational efficiency via better memory locality, whereas smaller units allow finer-grained, more precise selection of important information.

\textbf{Block}-based methods select blocks of units to approximate full attention. For prefilling, Star Attention approximates attention using local blocks and the first prefix block. MInference's Block-Sparse pattern \citep{Jiang2024MInference1A} additionally incorporates a set of dynamically selected blocks for each chunk of query tokens. For decoding, Quest \citep{Tang2024QuestQS} and InfLLM \citep{xiaoinfllm} divide the KV cache into contiguous pages and select a subset of them for each decoded token. 

\textbf{Vertical--slash} patterns represent another essential class of units. Early sparse attention methods like LM-Infinite \citep{Han2023LMInfiniteZE} and StreamingLLM \citep{Xiao2023EfficientSL} utilised local sliding windows supplemented by prefix tokens shared globally, also known as attention sinks. Extending this approach, Tri-shape \citep{Li2024SCBenchAK} added full attention for suffix tokens, whereas SnapKV \citep{Li2024SnapKVLK} introduced dynamically chosen vertical columns. MInference \citep{Jiang2024MInference1A} built on this by adding diagonal slashes at arbitrary offsets beyond the local window.\footnote{Interestingly, to efficiently compute attention along these diagonals, MInference uses 64×64 blocks aligned with these diagonals rather than computing attention for individual query--key pairs.} \Cref{fig:sparse_attention_patterns} illustrates the main attention patterns covered in this section.

\subsection{Importance Estimation}

To identify which specific units to retain, one can use fixed patterns---applied identically across all inputs---or dynamic patterns that adapt to the content being processed. Fixed patterns introduce no computational overhead but cannot adapt to varying input requirements, while dynamic patterns better preserve model quality but require additional computation to identify important connections.

\textbf{Fixed} patterns are identified with offline calibration to work well across all inputs. StreamingLLM \citep{Xiao2023EfficientSL}, LM-Infinite \citep{Han2023LMInfiniteZE} and MoA \citep{Fu2024MoAMO} determine the number of initial tokens (attention sinks) and the width of a local sliding window.

\textbf{Content-aware} methods typically estimate the importance of QK units (tokens, blocks, or diagonals) to retain only the top-k most relevant ones, maximising attention score recall. They use lightweight heuristics such as approximated attention scores from highest-magnitude dimensions \citep[SparQ;][]{Ribar2023SparQAB} or block-wise pooled token representations \citep{Jiang2024MInference1A}. Some approaches subsample queries \citep[SampleAttention;][]{Zhu2024SampleAttentionNA}, recognising that recent query tokens often provide better indicators of KV unit importance, as in MInference's Vertical-Slash \citep{Jiang2024MInference1A} and SnapKV \citep{Li2024SnapKVLK}. During decoding, aggregated attention scores \citep[H2O;][]{Zhang2023H2OHO} or the latest query \citep[TOVA;][]{Oren2024TransformersAM} guide the selection of KV units, again prioritising units likely to receive high attention weights. Some methods incorporate complementary heuristics alongside attention scores, such as norms of keys \citep{Devoto2024ASA} or values \citep{Guo2024AttentionSI}.

Critically, the cost of sparse attention includes both the sparse operation and the importance estimation overhead, and the realised speed-up depends not only on the FLOPs saved but also on how efficiently a kernel can exploit the resulting sparsity pattern. During prefilling, exact per-query importance estimation is quadratic, and fine-grained (token-to-token or token-to-page) selection yields irregular memory access patterns that are poorly matched to block-based FlashAttention kernels. Training-based DeepSeek Sparse Attention \citep[DSA;][]{deepseekai2025deepseekv3} sidesteps both constraints by learning a ``lightning indexer'' whose scoring remains quadratic in sequence length but operates in a much smaller head dimension than full attention---so the $O(L^2 d_\text{head})$ cost is reduced enough to be practical---and by shipping a purpose-built sparse kernel that exploits the resulting token-level sparsity. In the training-free regime we study, no method to date has combined an efficient fine-grained estimator with a kernel that turns the resulting sparsity into wall-clock gains, so practical methods either select fine-grained units globally (e.g., vertical columns shared across all queries) or use coarser block-to-block selection. During decoding, per-query selection is feasible since only one query is processed per step, enabling methods like Quest to perform finer token-to-page selection and tolerate higher compression (\cref{sec:results_individual}).

\subsection{Budget Allocation}

The third dimension in sparse attention design is budget allocation: distributing computational resources across model components (layers and heads) for a target sparsity. This involves a trade-off between uniform simplicity and adaptive expressivity.

\textbf{Uniform} allocation assigns an equal budget (tokens or blocks) to each head as in Block-Sparse \citep{Jiang2024MInference1A} and SnapKV \citep{Li2024SnapKVLK}. This is computationally simple but overlooks that layers and heads contribute differently to accuracy and have diverse attention sparsity \citep{Zhang2024UnifyingKC}.

\textbf{Adaptive} methods vary budget allocation. PyramidKV \citep{Cai2024PyramidKVDK} and PyramidInfer \citep{Yang2024PyramidInferPK} observe that attention score entropy decreases with layer depth, allocating larger budgets to early layers. Mixture of Sparse Attention \citep[MoA;][]{Fu2024MoAMO} uses Taylor approximations to optimally distribute the global budget across layers. Within layers, Ada-KV \citep{Feng2024AdaKVOK} flexibly allocates by selecting top-$(k \times h)$ tokens (where $h$ is head count), allowing critical heads to retain more keys while pruning others. \textit{Threshold-based} allocation offers maximum flexibility by removing a fixed global budget. Methods like Twilight \citep{Lin2025TwilightAA}, FlexPrefill \citep{lai2025flexprefill}, Tactic \citep{Zhu2025TacticAS}, and SampleAttention \citep{Zhu2024SampleAttentionNA} set coverage thresholds (e.g., 95\% of attention mass). Each head dynamically selects units to meet these thresholds, allowing high-entropy attention heads to consume more budget and the overall budget to vary per sample.

\begin{table*}[!t]
\centering
\resizebox{\textwidth}{!}{
    \begin{tabular}{ll|llll}
    \toprule
    & Method & Unit & Budget & KV Cache Management \\
    \midrule
    \multirow{3}{1em}{\rotatebox[origin=c]{90}{Prefill}} & \textbf{Vertical-Slash} \citep{Jiang2024MInference1A}     & verticals and slashes & uniform & N/A \\
    & \textbf{FlexPrefill} \citep{lai2025flexprefill}    & verticals and slashes & threshold-based & N/A \\
    & \textbf{Block-Sparse} \citep{Jiang2024MInference1A} & blocks & uniform & N/A \\
    \midrule
    \multirow{3}{1em}{\rotatebox[origin=c]{90}{Decode}} & \textbf{SnapKV} \citep{Li2024SnapKVLK} & tokens &  uniform & eviction \\
    & \textbf{Ada-SnapKV} \citep{Feng2024AdaKVOK}  & tokens & adaptive & eviction \\
    & \textbf{Quest} \citep{Tang2024QuestQS} & pages & uniform & full cache \\
    \bottomrule
    \end{tabular}
}
\caption{Full list of content-aware sparse attention methods benchmarked in our experiments. These represent diverse strategies in terms of units, budget allocation, and KV cache management.}
\vspace{-5mm}
\label{tab:sa_methods}
\end{table*}

\subsection{KV Cache Management}

The final dimension distinguishes methods based on KV cache management during decoding.

\textbf{KV cache eviction} methods (e.g., H2O \citep{Zhang2023H2OHO}, SnapKV \citep{Li2024SnapKVLK}) permanently discard selected tokens based on estimated importance, directly reducing memory footprint but sacrificing information fidelity as discarded tokens cannot be recovered.

\textbf{Full KV cache} retention methods (e.g., Quest \citep{Tang2024QuestQS}, SparQ \citep{Ribar2023SparQAB}) maintain the entire cache but optimize computation by selectively loading only necessary KV pairs during attention calculation. While incurring small memory overhead for auxiliary data structures needed for importance estimation, they avoid information loss and can operate effectively at higher sparsity levels compared to eviction-based methods, though they do not reduce peak memory requirements.

\section{Experimental Setup}
\label{sec:exp_setup}

\subsection{Models}
We perform experiments primarily on Qwen 2.5 \citep{qwen2} (7B, 14B, 32B, 72B parameters), complemented by Llama 3.1 \citep{Dubey2024TheL3} (8B, 70B) and Gemma 3 \citep{gemma3} (4B, 12B, 27B). All three families use instruction-tuned variants to support chain-of-thought evaluation. Qwen 2.5 was selected as our primary family as it uniquely satisfies strict methodological requirements for controlled scaling experiments—see \cref{sec:model_details} for rationale. For Qwen and Llama, we modify the attention mechanism across all layers. Gemma 3 employs hybrid attention where 5 out of 6 layers use sliding window attention (1024 tokens) by design; we apply sparse attention methods only to the remaining dense (global attention) layers. We preserve the original architectures and utilise the vLLM inference engine \citep{kwon2023efficient} with full bf16 precision. Implementation details are in \cref{sec:implementation_details}.

\subsection{Sparse Attention Methods}
We evaluate six state-of-the-art sparse attention methods (\cref{tab:sa_methods}), which we choose as a representative set spanning across the key dimensions described in Section \ref{sec:background}. We focus exclusively on content-aware methods, as prior work has demonstrated that fixed patterns consistently underperform their dynamic counterparts \citep{Li2024SCBenchAK}.

\begin{table*}[!t]
\centering
\footnotesize
\begin{tabularx}{\textwidth}{lXccc}
\toprule
\textbf{Task Name} & \textbf{Description} & \textbf{Dispersion} & \textbf{Scope} & \textbf{Natural} \\
\midrule
    QA (SQuAD) & Open-ended QA on a specified document among distractors & Low & Low & \checkmark \\
    QA (QuALITY, TOEFL) & Multiple-choice QA on a specified document among distractors & Low & Low & \checkmark \\
    Ruler NIAH & Extract 4 values for specified keys among many distractor key-value pairs & Low & Low & \texttimes \\
    Ruler VT & Identify variables that resolve to a specific value via chained assignments & High & Low & \texttimes \\
    Ruler CWE & Identify the 10 most frequent words from a list with distractors & Low & High & \texttimes \\
    Story Retrieval & Answer 16 factoid-style questions about specific chapters in a long narrative & Low & Low & \checkmark \\
    Story Multi-hop & Identify the item acquired immediately before a target item across chapters & High & Low & \checkmark \\
    Story Filtering & Identify chapters where no item purchases occurred in a long narrative & Low & High & \checkmark \\
\bottomrule
\end{tabularx}
\caption{Summary of 9 evaluation tasks: QA tasks are based on existing datasets---SQuAD \citep{Rajpurkar2018KnowWY}, QuALITY \citep{Pang2021QuALITYQA}, TOEFL \citep{tseng2016towards}---while NIAH, VT, and CWE are taken from the RULER benchmark \cite{Hsieh2024RULERWT}. The remaining three (Story Retrieval, Multi-hop, and Filtering) are our contribution: we automatically generate multi-chapter narratives to evaluate the same skills as RULER tasks but expressed in naturalistic text. For each task, we indicate whether it has High or Low \emph{dispersion} (information is difficult to locate), High or Low \emph{scope} (large amount of necessary information), and whether it is based on \emph{natural} text or is synthetic.}
\label{tab:task_summary}
\end{table*}

\subsection{Tasks}
\label{ssec:tasks}
We evaluate 9 diverse tasks selected to reflect different characteristics along 3 key dimensions known to influence sparse attention performance: task difficulty---defined by \textit{Dispersion} (how hard it is to locate necessary information) and \textit{Scope} (how much information must be processed) \citep{Goldman2024IsIR}---and data \textit{Naturalness} (natural language vs.\ synthetic data). This multi-dimensional approach is motivated by recent findings that attention patterns vary significantly across task types: retrieval tasks often exhibit localised attention, while reasoning tasks show more uniform distributions that are challenging for sparse methods \citep{Liu2025CanLM, Chen2024MagicPIGLS, Li2024SCBenchAK}. The naturalness dimension is also crucial, as synthetic tasks yield different token representation distributions compared to natural language \citep{Liu2024ReAttentionTI}. Our task suite therefore incorporates four core tasks from the \textbf{RULER} benchmark \citep{Hsieh2024RULERWT}—Retrieval (NIAH), Multi-hop reasoning (VT), Aggregation (CWE), and QA (SQuAD)—to provide controlled environments (mostly synthetic) for specific capabilities. We complement these with natural texts from benchmarks with minimal contamination risk \citep{Li2024LongCV}, such as QuALITY and TOEFL, though these represent low-dispersion, low-scope tasks. Thus, we additionally introduce three novel tasks (\textbf{Story} Retrieval, Multi-hop, Filtering) that translate RULER's challenging tasks (with high dispersion or scope) into naturalistic narratives, more representative of real-world use. We deliberately avoid open-ended tasks like summarisation due to unreliable evaluation metrics \citep{Yen2024HELMETHT, ye2024justice}, focusing instead on structured-output tasks requiring factual answers, enabling precise evaluation via Exact Match Accuracy, Intersection-over-Union (IoU), and F1 score (all ranging from 0 to 1). These tasks are summarised in \cref{tab:task_summary}, with detailed descriptions in \cref{sec:task_details} and examples in \cref{sec:example_task_inputs}.

\subsection{Evaluation Settings}
Our evaluation covers input lengths of 16k, 32k, and 64k tokens for all model families, with 128k evaluations limited to Qwen and Llama using Vertical-Slash and Quest only; Gemma exhibited near-zero performance at 128k. We use 100 samples per configuration for Qwen and 50 for Llama and Gemma. We evaluate all combinations of task, model size, sequence length, and sparse attention pattern at sparsity levels from 0 (dense) to 0.95 (i.e., $1/20$ attention budget), interpolating performance at intermediate points. We ensure input samples are within 95--100\% of the target maximum token length, providing a consistent basis for evaluating the impact of sequence length on performance. Following \citet{Karpinska2024OneTA}, we adopt a structured prompt format that encourages models to explicitly reason through chain-of-thought before providing answers in a consistent, parsable structure (see \cref{sec:prompt_template}). As metrics of computational cost, we report FLOPS for prefilling and memory access for decoding, as these reflect the respective computational bottlenecks of each phase (see \cref{sec:background}). \Cref{sec:cost_breakdown} provides more details, including indexing costs for sparse attention methods.

\section{Results}
\label{sec:results}

\subsection{\isoCost Analysis}
\label{sec:results_isoflop}

\begin{figure*}[t!]
    \centering
    \includegraphics[width=\linewidth]{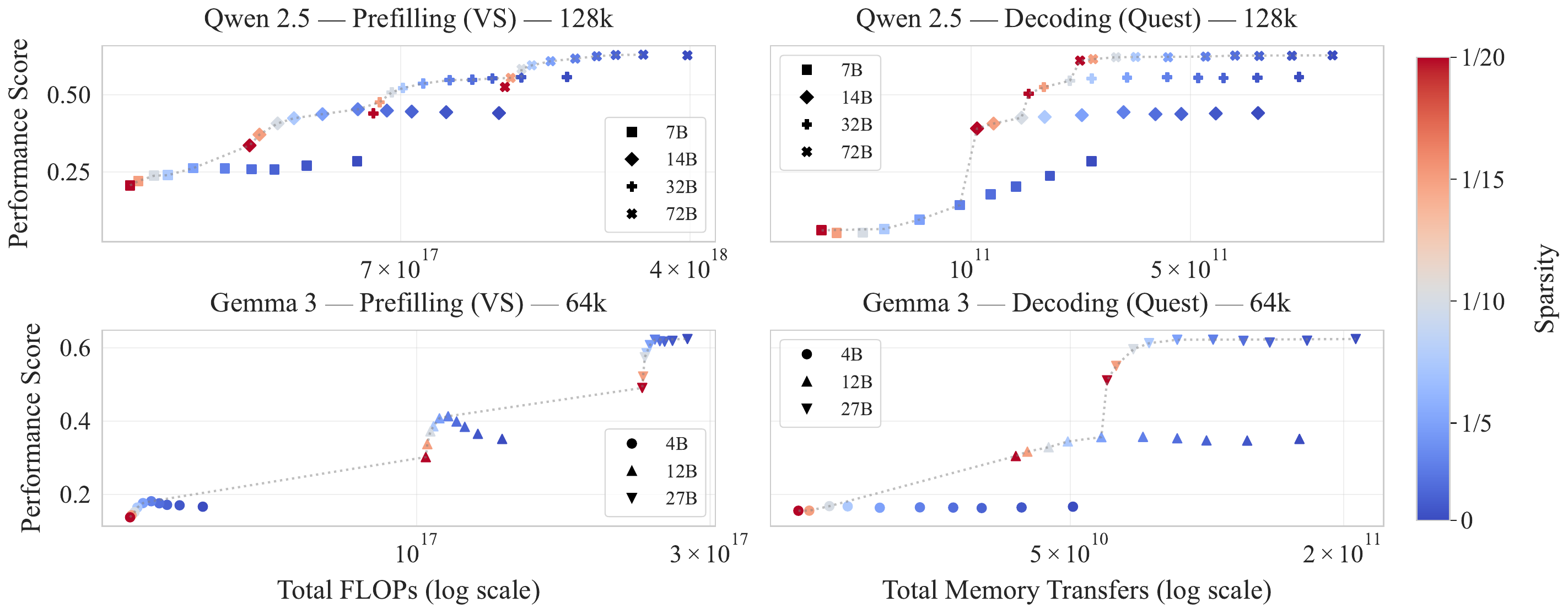}
    \caption{\isoCost analysis for Qwen 2.5 (128k tokens) and Gemma 3 (64k tokens). Each point corresponds to a (model size, sparsity) configuration, with performance aggregated across 9 tasks. \textbf{Left column}: prefilling with Vertical-Slash \citep{Jiang2024MInference1A} (FLOPs). \textbf{Right column}: decoding with Quest \citep{Tang2024QuestQS} (memory transfers). Both costs are computed at batch size $B=64$ (see \cref{sec:cost_breakdown}). Standard error is negligible (\cref{app:statistical_error}) and omitted for visual clarity. Dotted lines show Pareto frontiers connecting configurations that are not dominated by any other configuration. Key findings: (1) sparsification enables larger sparse models to outperform smaller dense models at equivalent cost; (2) the impact of sparsity is less pronounced for Gemma due to its sliding-window architecture (\cref{fig:gemma_vs_qwen_comparison}).}
    \label{fig:isocost_analysis}
\end{figure*}

\textit{RQ1: Is sparse attention effective?}
Results in \cref{fig:isocost_analysis} illustrate the average performance across tasks against computational cost for different model sizes and levels of sparsity.\footnote{We approach this question using Vertical-Slash for prefilling and Quest for decoding, as these are, on average, the best-performing patterns for their respective inference phases (see \cref{sec:results_individual}).} As implementation-agnostic proxies for computational cost, we use FLOPs for prefilling and memory transfers for decoding, which correlate with wall-clock time under optimised implementations (see \cref{sec:cost_breakdown} for cost formulas and breakdowns). We visualise Pareto frontiers to identify configurations that offer the best performance-cost trade-offs, i.e., those not dominated by any other configuration in terms of both cost and performance.

\paragraph{Sparse attention improves the Pareto frontier.} In \cref{fig:isocost_analysis}, the Pareto frontier reveals an efficiency crossover where sparsification enables larger sparse models to outperform smaller dense ones at equivalent computational cost. For Qwen at 128k tokens, only high-sparsity configurations lie on the Pareto frontier. During prefilling, models with sparsity 0.8--0.93 (i.e., $1/5$ to $1/15$ attention budget) remain optimal, while sparsity 0.95 ($1/20$ budget) falls below the optimal boundary. Decoding shows better resilience to high sparsity, with even 0.95 sparsity configurations being preferable to smaller dense models. For Gemma, we observe similar trends during decoding, but configuration overlap is absent for prefilling---this reflects Gemma's lower baseline attention ratio because of its sliding-window architecture (see \cref{sec:background,sec:cost_breakdown}). 

\subsection{Per-Task Analysis}
\label{sec:results_individual}

\begin{figure*}[t!]
    \centering
    \includegraphics[width=\linewidth]{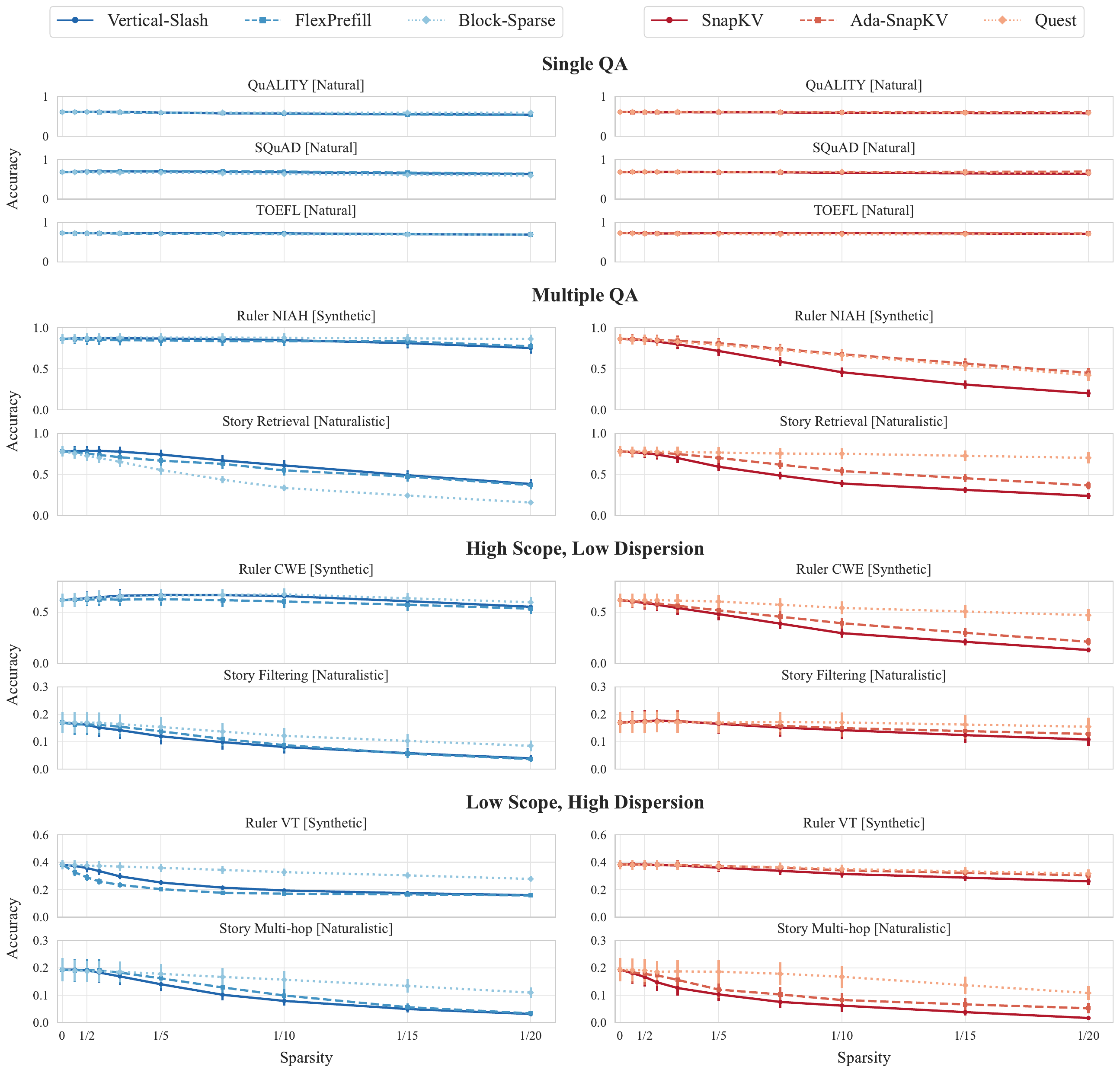}
    \caption{Per-task performance comparison of sparse attention methods, aggregated over Qwen 2.5, Llama 3.1, and Gemma 3 models at sequence lengths 16k, 32k, and 64k. Error bars indicate standard error. \textbf{Left column}: prefilling (Vertical-Slash, FlexPrefill, Block-Sparse). \textbf{Right column}: decoding (SnapKV, Ada-SnapKV, Quest). Tasks are grouped by information retrieval characteristics. Per-family breakdowns are provided in \cref{app:per_task}.}
    \label{fig:results_per_task}
\end{figure*}

\textit{RQ2: Which sparse attention method should practitioners use?}
\Cref{fig:results_per_task} presents per-task performance across sparse attention methods, aggregated over three model families and sequence lengths up to 64k. The 9 tasks introduced in \cref{ssec:tasks} are grouped by their information retrieval characteristics: Single QA (one query, localised answer), Multiple QA (multiple queries targeting distinct facts), High Scope/Low Dispersion (broad context, concentrated answers), and Low Scope/High Dispersion (narrow focus, scattered information). Three findings emerge from this analysis.

\paragraph{Prefill and decoding phases display different flexibility.} As discussed in \cref{sec:background}, the computational constraints of each inference phase---both the cost of importance estimation and the ability of a kernel to translate sparsity into wall-clock savings---shape which patterns are practical, which in turn affects generalisation across tasks. In the training-free regime, no published method combines a sub-quadratic (or approximate-quadratic) fine-grained estimator with a kernel that effectively exploits the resulting irregular sparsity during prefilling; the methods we evaluate therefore fall into one of two strategies: global selection of fine-grained units (Vertical-Slash) or block-to-block selection (Block-Sparse). Neither strategy dominates---the optimal choice is task-dependent.

During prefilling, Vertical-Slash shows strong performance on retrieval tasks (Low Scope, Low Dispersion) by enabling fine-grained token selection for locating specific facts. Tasks demanding broader context access or multi-step reasoning (High Scope or Dispersion, e.g., Ruler VT, Story Filtering) benefit from Block-Sparse, which selects distinct key-token blocks for each query block, accommodating the processing of multiple independent segments. 

During decoding, token-to-page selection becomes cheap since only one query is processed per step. This greater flexibility enables Quest to generalise better across tasks and tolerate higher compression than either prefilling approach, while retaining the full KV cache. Eviction-based decoding methods (SnapKV, Ada-SnapKV) that permanently discard tokens illustrate the cost of sacrificing the full cache---irreversible compression is detrimental when discarded tokens become relevant later, though this comes with the benefit of reduced memory footprint. Nevertheless, Quest can degrade on synthetic tasks such as Ruler NIAH, where random symbol sequences yield less distinguishable key representations compared to natural language \citep{Liu2024ReAttentionTI}---Quest's page-level granularity amplifies this effect, as coarser blocks struggle more than Ada-SnapKV's token-level selection to differentiate between unrelated token sets.

\paragraph{Dynamic budget allocation benefits are phase-dependent.} Adaptive methods that allocate different budgets across layers or sequences yield inconsistent results. During prefilling, FlexPrefill matches or underperforms Vertical-Slash's uniform allocation, likely due to the ``attention sink phenomenon''~\citep{Chen2024MagicPIGLS}: threshold-based selection captures high-attention tokens while missing information in the distribution's long tail. During decoding, Ada-SnapKV consistently outperforms uniform SnapKV, particularly on multi-query tasks (Story Retrieval), though both eviction methods remain inferior to Quest's full-cache approach.

\paragraph{Sparsity tolerance varies dramatically across tasks.} The gap between task groups reveals a deployment risk: methods achieving high sparsity on easy tasks may fail on harder ones. Single QA tasks (QuALITY, SQuAD, TOEFL) tolerate sparsity 0.95 ($1/20$ budget) with minimal degradation across all methods. Multiple QA tasks (Ruler NIAH, Story Retrieval) show substantial degradation at sparsity 0.8--0.9 ($1/5$ to $1/10$ budget). Tasks with High Scope or High Dispersion degrade even at modest sparsity (0.5--0.67, i.e., $1/2$ to $1/3$ budget) for some methods. Evaluating sparse attention only on Single QA benchmarks---or averaging across task types---masks these vulnerabilities. Robust deployment requires testing across diverse task characteristics, as sparsity levels safe for retrieval tasks can cause failures on aggregation or multi-hop reasoning. Moreover, sequence naturalness affects methods asymmetrically---Quest outperforms Ada-SnapKV on natural-language retrieval (Story Retrieval) but underperforms on synthetic retrieval (Ruler NIAH)---underscoring the need for benchmarks spanning both natural and synthetic data.

\subsection{Sequence Length Effects}
\label{sec:results_seq_length}

\begin{figure}[t!]
    \centering
    \includegraphics[width=\linewidth]{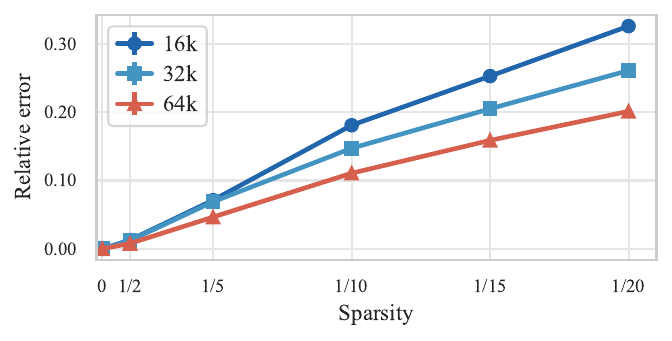}
    \caption{Sequence length effects on sparsity tolerance. Relative error is $(\bar{p}_{\text{dense}} - \bar{p}_{\text{sparse}}) / \bar{p}_{\text{dense}}$, where $\bar{p}$ denotes mean performance. Results aggregated across all tasks, methods, and models (Qwen 2.5, Llama 3.1, Gemma 3). Per-family breakdowns are provided in \cref{app:seq_length_absolute}.}
    \label{fig:seq_length_relative}
\end{figure}

\textit{RQ3: How does sequence length affect tolerance to sparse attention?} \Cref{fig:seq_length_relative} shows that \textbf{for a fixed attention budget fraction, longer sequences incur smaller degradation}: for example, at a $1/20$ budget, the relative error decreases from $\approx 0.33$ (16k) to $\approx 0.26$ (32k) and $\approx 0.20$ (64k). This indicates that the same sparsity ratio becomes less harmful as the sequence length grows. This pattern holds consistently across all model families. \citet{Nawrot2024DynamicMC} observe similar results for a training-aware KV compression method: their learned mechanism applies lower sparsity at the beginning of sequences and increases sparsity with sequence length. This behaviour may be explained by Herdan's law \citep{herdan1960type}, which posits that new information becomes rarer over time, facilitating higher sparsity with distance.

To relate this trend to \textbf{budget scaling}, we interpret the plot as approximate \emph{iso-error} curves. For a target relative error of $\approx 0.2$, the required budget fractions are roughly $1/10$ (16k), $1/15$ (32k), and $1/20$ (64k). In contrast, a \textbf{fixed token budget} would imply fractions $1/10 \rightarrow 1/20 \rightarrow 1/40$ as length grows, which already exceeds the $\approx 0.2$ error target at 32k (the $1/20$ point is $\approx 0.26$). For a stricter target such as $\approx 0.1$, the scaling is less uniform: $1/5$ stays below $0.1$ for both 16k and 32k, while 64k requires only a modest reduction in sparsity to stay near $0.1$. These observations imply that the optimal token budget should grow \emph{sublinearly} with sequence length: doubling the context does not require doubling the token budget, but keeping the budget constant would incur increasing degradation. While current dynamic methods lack robustness (\cref{sec:results_individual}), developing reliable sublinear budget allocation mechanisms remains a promising direction for future work.

\section{Conclusions}
\label{sec:conclusions}

This study provides the largest-scale empirical analysis of training-free sparse attention to date, covering three model families (Qwen 2.5, Llama 3.1, Gemma 3), model scales (4B--72B parameters), sequence lengths (16K--128K tokens), sparsity levels up to 0.95 (i.e., $1/20$ attention budget), and nine diverse long-sequence tasks. We organise the rapidly evolving landscape of sparse attention methods into a taxonomy along four design axes and introduce novel benchmarks consisting of natural texts that are fully controllable yet challenging. Our analysis yields three key insights.

\noindent\textbf{Evidence of effectiveness.} Sparse attention enables larger models to outperform smaller dense ones at equivalent computational cost, improving the Pareto frontier. Thus, sparsity becomes crucial for optimal LLM scaling.

\noindent\textbf{Practical deployment guidance.} Method selection should be task-aware: fine-grained token selection (e.g., Vertical-Slash) excels at retrieval, chunk-based methods (e.g., Block-Sparse) suit reasoning and aggregation, and Quest provides robust decoding across most scenarios.

\noindent\textbf{Design recommendations.} Longer sequences tolerate higher sparsity while maintaining accuracy. This suggests that fixed-budget methods deployed in production are suboptimal; future designs should adapt sparsity levels to sequence length, possibly growing the token budget sublinearly.

\section*{Limitations}

First, we evaluate only training-free sparse attention methods. Training-based approaches could reduce train-inference mismatch, but require substantial computational resources and access to proprietary training data.

Second, our experimental coverage, while extensive, is bounded. We evaluate three model families (Qwen 2.5, Llama 3.1, Gemma 3) that met our methodological requirements for controlled scaling experiments with native long-context support; other families may exhibit different behaviour. We test only instruction-tuned models; reasoning models with extended chain-of-thought capabilities (e.g., o1, DeepSeek-R1) may have different attention patterns and sparsity tolerance. Our nine tasks, though selected to span diverse dispersion levels, processing scopes, and data naturalness, do not exhaustively cover all long-context scenarios—open-ended tasks like summarisation were excluded due to unreliable automated metrics. Additionally, experiments at 128k tokens are limited due to low baseline performance and lack of robustness across models; more conclusive evidence on how sequence length affects sparse attention scaling requires stronger long-context models.

Third, we report hardware-agnostic computational costs (FLOPs and memory access) rather than wall-clock timings. Actual speedups depend on hardware, batch size, and implementation quality, which vary across deployment environments.

Fourth, we do not investigate interactions between sparse attention and other model efficiency techniques such as quantisation, weight pruning, or mixture-of-experts sparsity. These methods are often combined in practice, and their joint effects on attention sparsity tolerance remain unexplored.

Fifth, our analysis treats prefill and decoding sparsity independently, but end-to-end efficiency depends on their joint allocation. For long-input, short-output workloads (as in our evaluation) prefill attention dominates total compute, so even modest prefill sparsity can yield larger wall-clock gains than aggressive decode compression; long-output regimes (e.g., reasoning models) invert this balance. Jointly optimising prefill and decoding sparsity as a function of the input/output length ratio---with cumulative memory reads as a unified implementation-agnostic cost proxy across phases---is an important direction for future work.

Sixth, the methods we study estimate importance from attention scores, implicitly targeting attention-recall maximisation rather than downstream task performance. How close a recall-maximising mask is to a task-loss-optimal one remains unknown; measuring this gap---ideally while ignoring estimation and kernel-efficiency constraints to isolate the pattern itself---would quantify how much headroom remains beyond better estimators and could motivate alternative selection criteria.

\section*{Acknowledgements}
This work is supported by the ERC Starting Grant AToM-FM (101222956), awarded to Edoardo M. Ponti, and by the UKRI Centre for Doctoral Training in Natural Language Processing, funded by the UKRI (grant EP/S022481/1) and the University of Edinburgh, School of Informatics and School of Philosophy, Psychology \& Language Sciences.

\bibliography{utils/custom}
\bibliographystyle{utils/acl_natbib}

\clearpage
\appendix

\section{Experimental details}
\label{sec:exp_details}

\subsection{Implementation Details}
\label{sec:implementation_details}

This section provides supplementary details on the sparse attention patterns evaluated, focusing on hyperparameter tuning and specific configurations used to achieve target sparsity levels. We tuned hyperparameters for each pattern using ablation studies on the Qwen-7B model with a 16K sequence length across all tasks, varying sparsity from 0 to 0.9. Our main experiments evaluated performance at sparsity levels 0.33, 0.5, 0.6, 0.7, 0.8, 0.87, 0.9, 0.93, 0.95 (corresponding to attention budgets $1/1.5$, $1/2$, $1/2.5$, $1/3.33$, $1/5$, $1/7.5$, $1/10$, $1/15$, $1/20$), using linear interpolation for intermediate values where necessary. Table~\ref{tab:pattern_parameters} summarizes the final parameters we used for each pattern, sequence length, and sparsity level. In total we evaluated 7065 configurations with 100 samples per configuration. We used approximately 4 compute nodes with 8 H100 GPUs each for 21 days.

\subsubsection{Block-Sparse Attention}
\begin{figure}[ht]
\centering
\includegraphics[width=\linewidth]{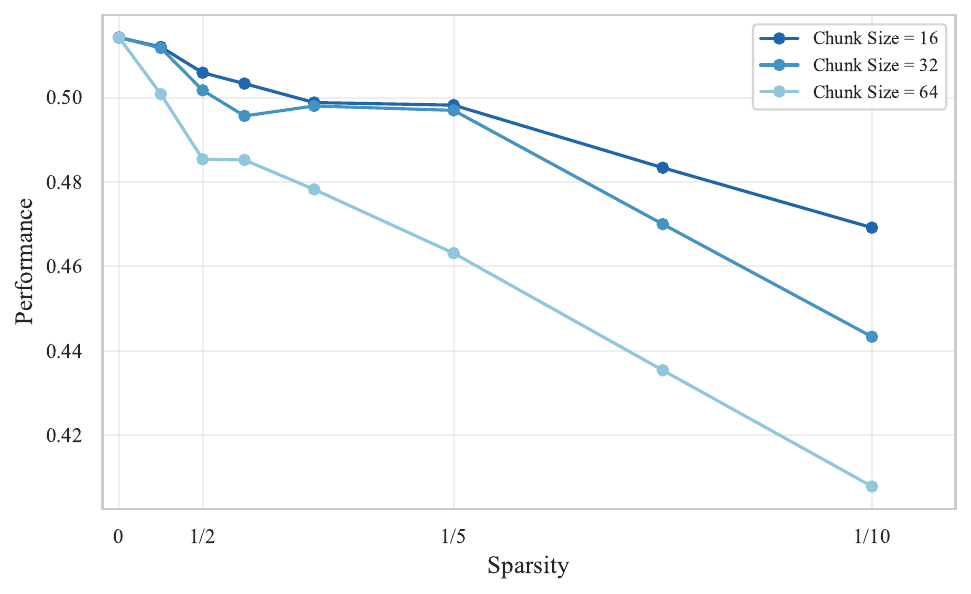}
\caption{Block-Sparse block size.}
\label{fig:block_size_ablation}
\end{figure}

We implement block-sparse attention by dividing the attention matrix into fixed-size blocks. The original implementation is available under the MIT license. Based on our ablation studies (Figure~\ref{fig:block_size_ablation}), we selected a block size of 16x16, as smaller blocks consistently yielded better performance. To achieve a target sparsity level, we select the top-$k$ key blocks for each query block, where $k$ is determined via binary search. We always preserve attention sinks (the first key block) and local context (diagonal key blocks corresponding to the query block).

\subsubsection{Vertical-Slash Pattern}
\begin{figure*}[ht]
\centering
\includegraphics[width=\textwidth]{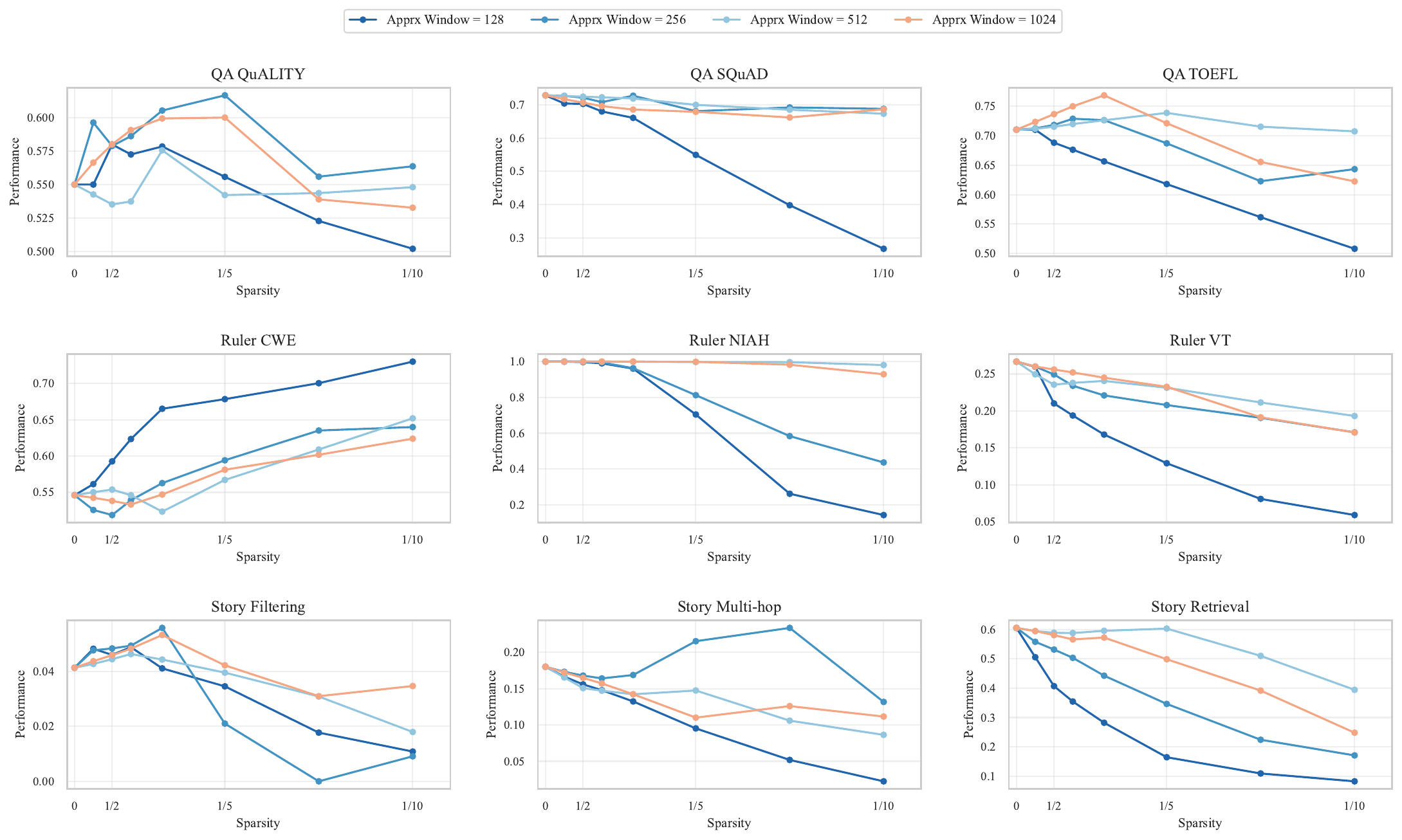}
\caption{Vertical-Slash approximation window ablation per task.}
\label{fig:vs_per_task}
\end{figure*}

We implement the Vertical-Slash pattern \citep{Jiang2024MInference1A}, available under the MIT license, by allocating a uniform budget to global (vertical columns) and local (slash diagonals) attention components. We select the most important verticals and slashes by approximating attention scores using a limited window of recent query tokens. Our ablation studies (Figure~\ref{fig:vs_per_task}) revealed task-dependent optimal approximation window sizes: 512 tokens for retrieval-heavy tasks (Ruler NIAH, Story Retrieval) and 256 tokens for other tasks. This observation correlates with the typical query lengths for these tasks (see Table~\ref{tab:question_token_stats}). We consistently preserve the first 4 (prefix) and the most recent 64 (local) tokens. To achieve target sparsity levels, we compute the required number of verticals and slashes based on collected attention statistics for each sequence length.

\begin{table}[ht]
    \centering
    \small
    \caption{Statistics of token lengths of question and instruction for each task across 100 samples, informing the choice of approximation window size for Vertical-Slash and FlexPrefill. The Min--Max column reports the observed range.}
    \label{tab:question_token_stats}
    \begin{tabular}{lcc}
    \toprule
    \textbf{Task} & \textbf{Mean} & \textbf{Min--Max} \\
    \midrule
    QA QuALITY & 243.63 & 196--423 \\
    QA SQuAD & 217.08 & 210--235 \\
    QA ToeflQA & 237.67 & 202--270 \\
    RULER CWE & 227.00 & 227--227 \\
    RULER NIAH & 337.74 & 330--350 \\
    RULER VT & 230.00 & 230--230 \\
    Story Filtering & 184.00 & 184--184 \\
    Story Multi-hop & 192.97 & 192--195 \\
    Story Retrieval & 457.54 & 452--462 \\
    \bottomrule
    \end{tabular}
\end{table}

\subsubsection{FlexPrefill}
\begin{figure}[ht]
\centering
\includegraphics[width=\linewidth]{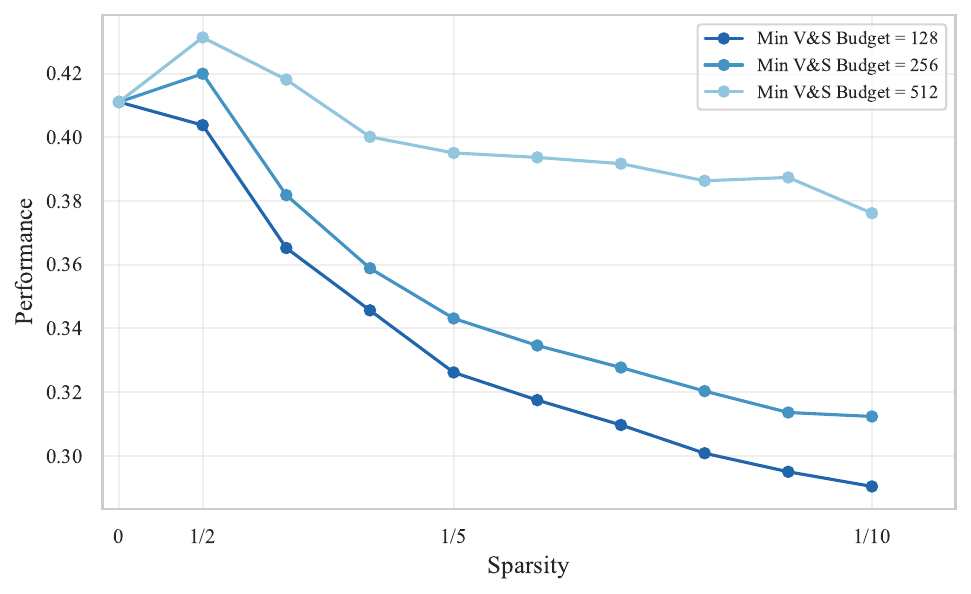}
\caption{FlexPrefill min budget.}
\label{fig:minbudget_flexprefill}
\end{figure}

We implement FlexPrefill \citep{lai2025flexprefill}, available under the Apache-2.0 license, which enhances Vertical-Slash by introducing dynamic budget allocation per layer and head, controlled by a coverage parameter $\alpha$ and a minimum budget (\texttt{min\_budget}). We set $\tau=0$ in our experiments, hence disabling Query-Aware attention. This choice stemmed from two key considerations: first, our preliminary tests indicated no significant performance gains from enabling it, aligning with the findings reported in the original work; second, this setting isolates the dynamic budget allocation mechanism, allowing us to specifically evaluate its impact compared to the fixed allocation used in the Vertical-Slash pattern. We employ the same task-dependent approximation windows (256 / 512 tokens) and critical token preservation strategy (first 4 prefix, most recent 64 local) as in our Vertical-Slash implementation. Our ablations (Figure~\ref{fig:minbudget_flexprefill}) indicated that setting \texttt{min\_budget} to 512 significantly improved performance, suggesting the importance of maintaining a minimum level of connectivity during prefilling. We achieved target compression ratios by selecting the appropriate $\alpha$ based on attention statistics while keeping \texttt{min\_budget} fixed at 512. For high compression ratios where dynamic allocation proved less effective, we set $\alpha=0$, effectively reverting to a uniform allocation of Vertical-Slash.

\subsubsection{SnapKV}
\begin{figure}[ht]
\centering
\includegraphics[width=\linewidth]{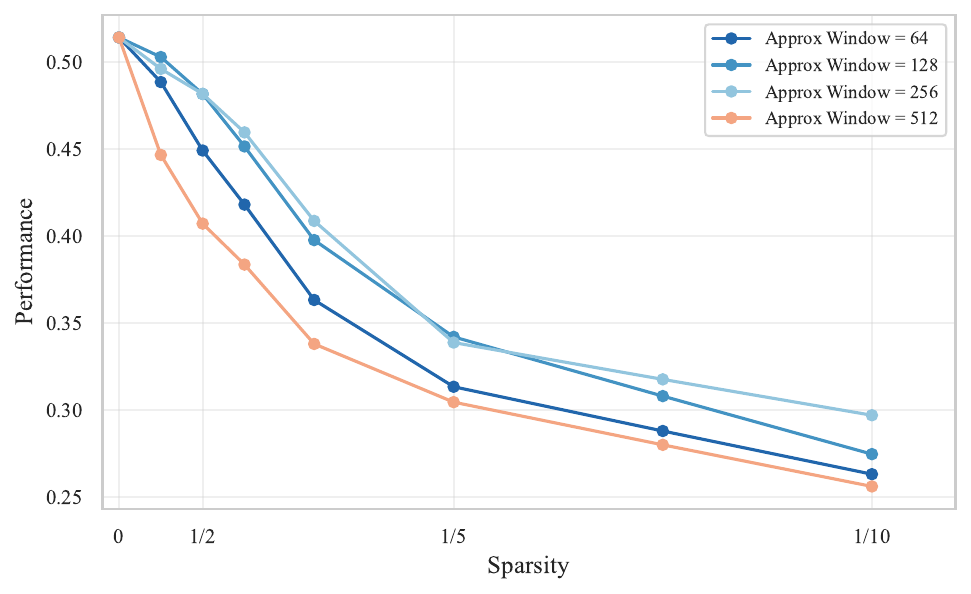}
\caption{SnapKV/Ada-SnapKV approximation window.}
\label{fig:snapkv_approx_window}
\end{figure}

\begin{figure}[ht]
\centering
\includegraphics[width=\linewidth]{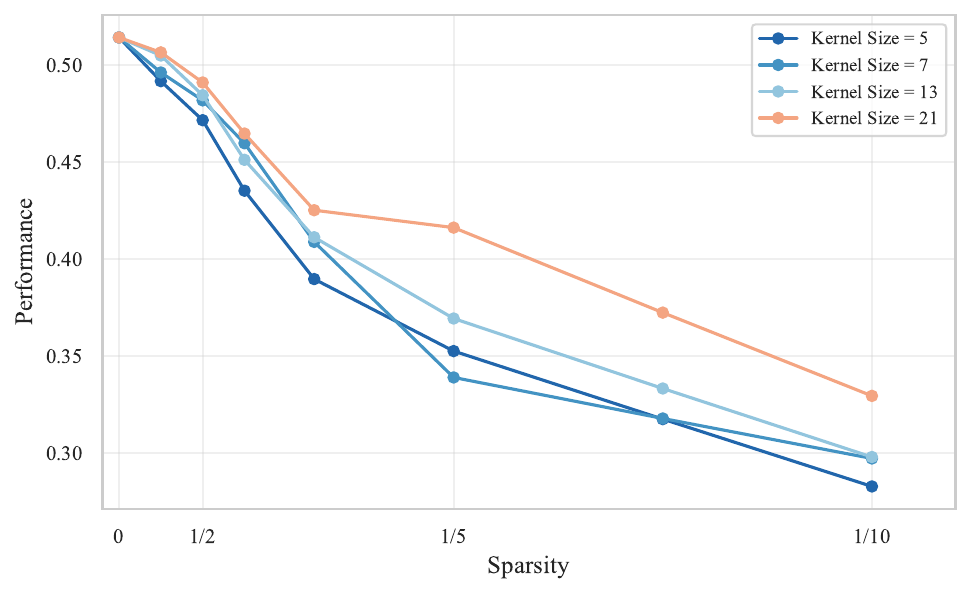}
\caption{SnapKV/Ada-SnapKV kernel size.}
\label{fig:snapkv_kernel_size}
\end{figure}

We implement SnapKV \citep{Li2024SnapKVLK}, available under the CC-BY 4.0 license, by compressing the Key-Value (KV) cache after the prefilling stage and applying a uniform token budget across all heads for the subsequent decoding phase. We predict token importance for decoding by computing attention scores using a window of recent query tokens (approximation window). Our ablations showed an optimal approximation window size of 256 tokens (Figure~\ref{fig:snapkv_approx_window}), with no significant task dependency observed, unlike Vertical-Slash and FlexPrefill. We smooth the calculated token importance scores using 1D average pooling with a kernel size of 21 (chosen based on Figure~\ref{fig:snapkv_kernel_size}). We always preserve the first 4 and the most recent 128 tokens. We control sparsity by setting the `token\_capacity` (token limit per head) to achieve the target sparsity level.

\subsubsection{Ada-SnapKV}
\begin{figure}[ht]
\centering
\includegraphics[width=\linewidth]{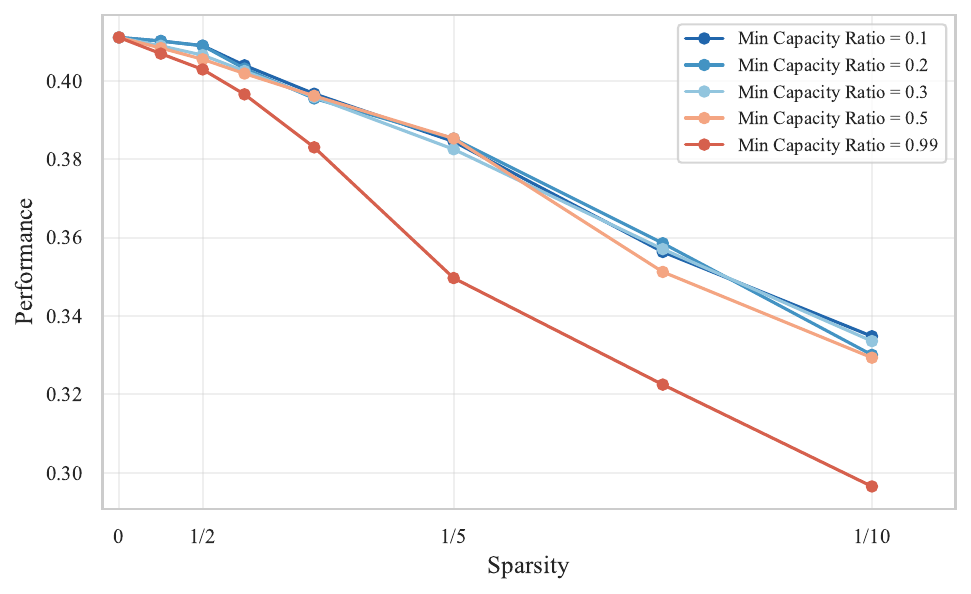}
\caption{Ada-SnapKV min budget.}
\label{fig:midbudget_ada_snapkv}
\end{figure}

We implement Ada-SnapKV \citep{Feng2024AdaKVOK}, available under the MIT license, which extends SnapKV by incorporating dynamic token budget allocation per head. One difference in our implementation between Ada-SnapKV and SnapKV is that we use max-aggregation (instead of averaging) across query positions and heads for score calculation; this empirically proved more effective for adaptive allocation but had no effect for uniform (SnapKV) allocation. We utilize the same smoothing kernel size (21) and critical token preservation strategy (first 4 prefix, most recent 128 local) as in our SnapKV implementation. Our ablations (Figure~\ref{fig:midbudget_ada_snapkv}) indicated that providing each head with a minimum budget of 20\% of its capacity was optimal. Performance was found to be less sensitive to minimum budget (performing well within the 10-50\% range) compared to FlexPrefill's sensitivity, but degraded sharply when approaching 100\% (uniform allocation), underscoring the benefits of dynamic allocation during decoding. We control sparsity by setting `token\_capacity`, identically to SnapKV.

\subsubsection{Quest}
\begin{figure}[ht]
\centering
\includegraphics[width=\linewidth]{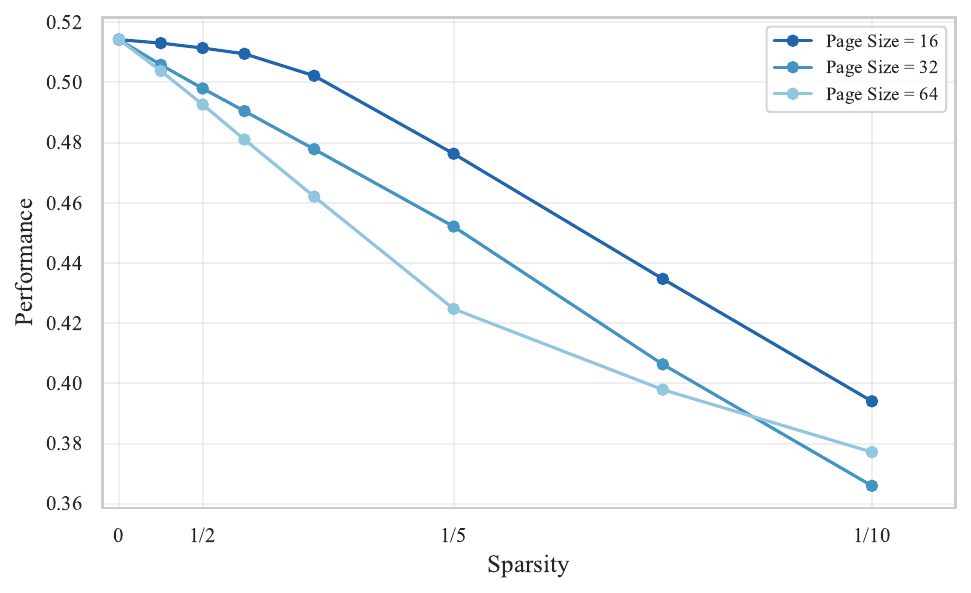}
\caption{Quest page size.}
\label{fig:quest_page_size_ablation}
\end{figure}

We implement Quest \citep{Tang2024QuestQS}, available under the CC-BY 4.0 license, which applies dynamic sparse attention during the decoding phase at the page level. Based on our ablations (Figure~\ref{fig:quest_page_size_ablation}), we used a page size of 16 tokens. We represent pages by their minimum and maximum key values to enable efficient similarity computation with queries. At each decoding step, we select the most relevant pages based on query-page similarity scores, always including the page containing the current token. We control sparsity by setting the `token\_budget` (number of tokens selected per step) to achieve the target sparsity level.

\begin{table*}[ht]
    \centering
    \caption{Pattern parameters for different sequence lengths and sparsity levels. At 128k tokens, we only evaluate Vertical-Slash and Quest.}
    \label{tab:pattern_parameters}
    \resizebox{\textwidth}{!}{%
    \begin{tabular}{llccccc}
    \toprule
    \textbf{Pattern} & \textbf{Parameter} & \textbf{Sequence Length} & \multicolumn{4}{c}{\textbf{Values for Different Sparsity Levels}} \\
    \midrule
    \multirow{3}{*}{Vertical \& Slash} & \multirow{3}{*}{Verticals/Slashes} & 16384 & \multicolumn{4}{c}{164, 240, 315, 400, 448, 576, 768, 1024, 1536, 2304} \\
    & & 32768 & \multicolumn{4}{c}{290, 384, 448, 576, 704, 1024, 1536, 2304, 3584, 4608} \\
    & & 65536 & \multicolumn{4}{c}{400, 448, 544, 640, 960, 1280, 2304, 4096, 6144, 8192} \\
    & & 128000 & \multicolumn{4}{c}{480, 768, 1024, 1536, 2048, 3584, 5632, 10240, 13312, 18432} \\
    \midrule
    \multirow{3}{*}{FlexPrefill} & \multirow{3}{*}{($\alpha$, min\_budget)} & 16384 & \multicolumn{4}{c}{(0, 164), (0, 240), (0, 315), (0, 400), (0.55, 512), (0.71, 512), (0.88, 512)} \\
    & & 32768 & \multicolumn{4}{c}{(0, 290), (0, 384), (0.45, 512), (0.6, 512), (0.7, 512), (0.8, 512), (0.92, 512)} \\
    & & 65536 & \multicolumn{4}{c}{(0, 400), (0.45, 512), (0.55, 512), (0.7, 512), (0.77, 512), (0.85, 512), (0.94, 512)} \\
    \midrule
    \multirow{3}{*}{Block Sparse} & \multirow{3}{*}{top\_chunks} & 16384 & \multicolumn{4}{c}{26, 35, 53, 71, 108, 188, 300} \\
    & & 32768 & \multicolumn{4}{c}{52, 69, 105, 141, 216, 376, 600} \\
    & & 65536 & \multicolumn{4}{c}{104, 139, 210, 283, 432, 752, 1200} \\
    \midrule
    \multirow{3}{*}{SnapKV/AdaSnapKV} & \multirow{3}{*}{token\_capacity} & 16384 & \multicolumn{4}{c}{819, 1092, 1638, 2183, 3276, 4915, 6553, 8192, 9830, 11468} \\
    & & 32768 & \multicolumn{4}{c}{1638, 2185, 3276, 4367, 6553, 9830, 13107, 16384, 19660, 22937} \\
    & & 65536 & \multicolumn{4}{c}{3276, 4371, 6553, 8735, 13107, 19660, 26214, 32768, 39321, 45875} \\
    \midrule
    \multirow{4}{*}{Quest} & \multirow{4}{*}{token\_budget} & 16384 & \multicolumn{4}{c}{816, 1088, 1632, 2176, 3280, 4912, 6560, 8192, 9824, 11472} \\
    & & 32768 & \multicolumn{4}{c}{1632, 2192, 3280, 4368, 6560, 9824, 13104, 16384, 19664, 22944} \\
    & & 65536 & \multicolumn{4}{c}{3280, 4368, 6560, 8736, 13104, 19664, 26208, 32768, 39328, 45872} \\
    & & 128000 & \multicolumn{4}{c}{6400, 8544, 12800, 17056, 25600, 38400, 51200, 64000, 76800, 89600} \\
    \bottomrule
    \end{tabular}
    }
\end{table*}

\subsection{Task Details}
\label{sec:task_details}

This section provides further details on the nine evaluation tasks used in our experiments. These tasks, summarized in \cref{tab:task_summary} at the end of this subsection, are grouped into Question Answering, synthetic tasks from RULER \citep{Hsieh2024RULERWT}, and our Story tasks. We specify key hyperparameters, evaluation metrics, and characterize each task along the axes of Scope (Low vs. High) and Dispersion (Low vs. High). Scope refers to the amount of information required, while Dispersion indicates how difficult it is to locate the relevant information within the context.

\subsubsection{Question Answering (QA)}
We use SQuAD \citep{Rajpurkar2018KnowWY} (CC BY-SA 4.0) from RULER (Apache-2.0) and two other QA datasets selected for minimal data contamination \citep{Li2024LongCV}: QuALITY \citep{Pang2021QuALITYQA} (CC BY 4.0) and ToeflQA \citep{tseng2016towards}\footnote{Originally released at \url{https://github.com/iamyuanchung/TOEFL-QA}. License information is missing; a GitHub issue requesting clarification was opened on August 7, 2023, but has not received a response.}.
\begin{itemize}
    \item \textbf{Setup:} Each example contains one answer-bearing document and distractor documents to reach the target sequence length. Documents are shuffled and numbered; the question refers to a specific document ID.
    \item \textbf{Preprocessing:} We remove duplicate question-context pairs and filter examples where the original context exceeds 8k tokens (ensuring space for at least one distractor at 16k sequence length).
    \item \textbf{Evaluation:} Exact Match Accuracy (QuALITY, ToeflQA multiple-choice), token-level F1 (SQuAD open-ended).
    \item \textbf{Characteristics:} Natural text. Requires identifying and processing a specific document, thus characterised by \textbf{Low Dispersion} and \textbf{Low Scope}. See \cref{example_task_input:qa}.
\end{itemize}

\subsubsection{Synthetic – RULER Tasks}
We use three synthetic tasks from the RULER benchmark \citep{Hsieh2024RULERWT} (Apache-2.0).
\begin{itemize}
    \item \textbf{Needle-in-a-Haystack (NIAH):} Extract values for 4 target keys from a document containing relevant and distractor key-value pairs (random hyphenated strings). Evaluated using Exact Match Accuracy. Requires finding specific items, characteristic of \textbf{Low Dispersion} and \textbf{Low Scope}. See \cref{example_task_input:niah}.
    \item \textbf{Common Word Extraction (CWE):} Identify the 10 most frequent words (appearing 30 times each) among distractor words (appearing 3 times each), sampled from a vocabulary of $\sim$9,000 English words\footnote{\url{https://github.com/mrmaxguns/wonderwordsmodule}}. Evaluated using Intersection-over-Union (IoU). Requires processing the entire context to count frequencies, demanding \textbf{Low Dispersion} (words are presented directly as a list, not obscured within complex structures) but \textbf{High Scope} (all words must be processed to determine frequencies). See \cref{example_task_input:cwe}.
    \item \textbf{Variable Tracking (VT):} Resolve variable assignments (direct or chained) to identify all variables matching a target value. Context includes repeated filler text (``\textit{The grass is green...}''). Evaluated using IoU. Requires tracking dependencies across the context, demanding \textbf{High Dispersion} (information location depends on chains) and \textbf{Low Scope} (only specific chains matter). See \cref{example_task_input:vt}.
\end{itemize}

\subsubsection{Semi-Synthetic – Story Tasks}
These tasks use procedurally generated multi-chapter narratives that scale with sequence length. Each chapter follows a schema involving travel, dialogue, and item transactions. See \cref{sec:example_story_narrative} for an example narrative. We release them under the CC BY 4.0 license.
\begin{itemize}
    \item \textbf{Story Retrieval:} Answer 16 factoid questions (e.g., location visited, item acquired) about specific chapters, with chapter IDs provided in the questions. Evaluated using Exact Match Accuracy. Requires accessing specific chapters, characteristic of \textbf{Low Dispersion} and \textbf{Low Scope}. See \cref{example_task_input:story_retrieval}.
    \item \textbf{Story Filtering:} Identify the three specific chapters where no item purchases occurred. The prompt explicitly asks for these three chapter IDs, and the narrative is constructed such that exactly three chapters meet this condition. Evaluated using IoU. Requires checking all chapters, demanding \textbf{Low Dispersion} (information is chapter-based) but \textbf{High Scope} (all chapters must be checked). We found this task to be challenging even for the largest models evaluated. See \cref{example_task_input:story_filtering}.
    \item \textbf{Story Multi-hop:} Given a target item, identify the item acquired immediately before it, requiring reasoning across the transaction history in multiple chapters. In our setup, an item is acquired in every chapter; this simplifies the task to locating the chapter where the target item was acquired and retrieving the item name from the immediately preceding chapter. We found this simplified version to be highly challenging, even for the largest models evaluated, thus we did not explore more complex variants (e.g., selective item acquisition requiring longer lookbacks). Evaluated using Exact Match Accuracy. Requires tracking history across the narrative, demanding \textbf{High Dispersion} (relevant transactions can be far apart) and \textbf{Low Scope} (only specific transaction pairs matter). See \cref{example_task_input:story_multihop}.
\end{itemize}

\subsection{Model Details}
\label{sec:model_details}

Our choice of Qwen 2.5 as the primary model family was driven by strict methodological requirements. We needed a model family satisfying three criteria simultaneously: (1) native 128k context support, since sparse attention benefits emerge primarily at very long sequences; (2) multiple model sizes maintaining reasonable (non-random) performance across all sequence lengths with consistent training procedures on identical data to enable rigorous size-based comparisons; and (3) instruction-tuned versions for chain-of-thought evaluation to mitigate short-output evaluation bias toward sparse decoding methods with dense prefill \citep{Yuan2024KVCC}.

After evaluating available open-source families, Qwen 2.5 was the only one meeting these requirements. Other families were excluded for the following reasons:
\begin{itemize}
    \item \textbf{Command}: Different training data across sizes (8B vs 32B/104B).
    \item \textbf{Llama 3.1/3.2}: Smaller models (1B, 3B) failed at 16k--32k sequence length on most tasks; 405B exceeded our computational budget.
    \item \textbf{Mistral}: Multiple fine-tunes but fewer than three sizes with consistent training.
    \item \textbf{Phi-3}: The 14B model showed unexpectedly poor long-context performance, worse than 4B according to RULER evaluations.
    \item \textbf{Yi}: Limited to 32k sequence length.
    \item \textbf{GPT-OSS}: Only two model sizes; additionally requires custom attention implementation with variable attention head biases, which is not supported by training-free sparse attention methods.
\end{itemize}

To broaden our scope and provide additional evidence for our findings, we also evaluate on Llama 3.1 (8B, 70B) and Gemma 3 (4B, 12B, 27B). Gemma 3 employs hybrid attention where 5 out of 6 layers use sliding window attention with a window size of 1024 tokens, while every 6th layer uses global (dense) attention. This makes Gemma 3 particularly interesting for our study: it is already heavily sparsified by design, and we apply training-free sparse attention methods only to the dense layers. This allows us to analyze whether additional sparsification benefits models that already incorporate architectural sparsity.

\begin{table}[ht]
    \centering
    \footnotesize
    \setlength{\tabcolsep}{4pt}
    \begin{tabular}{llcccl}
    \toprule
    \textbf{Family} & \textbf{Size} & \textbf{L} & \textbf{Q} & \textbf{KV} & \textbf{Huggingface} \\
    \midrule
    \multirow{2}{*}{Llama 3.1} & 8B & 32 & 32 & 8 & Llama-3.1-8B-Instruct \\
    & 70B & 80 & 64 & 8 & Llama-3.1-70B-Instruct \\
    \midrule
    \multirow{4}{*}{Qwen 2.5} & 7B & 28 & 28 & 4 & Qwen2.5-7B-Instruct \\
    & 14B & 48 & 40 & 8 & Qwen2.5-14B-Instruct \\
    & 32B & 64 & 40 & 8 & Qwen2.5-32B-Instruct \\
    & 72B & 80 & 64 & 8 & Qwen2.5-72B-Instruct \\
    \midrule
    \multirow{3}{*}{Gemma 3} & 4B & 34 & 8 & 4 & gemma-3-4b-it \\
    & 12B & 48 & 16 & 8 & gemma-3-12b-it \\
    & 27B & 62 & 32 & 16 & gemma-3-27b-it \\
    \bottomrule
    \end{tabular}
    \caption{Overview of models used in the evaluation. \textbf{L}: layers; \textbf{Q}/\textbf{KV}: number of query/key-value heads. Hugging Face IDs are prefixed with \texttt{meta-llama/}, \texttt{Qwen/}, and \texttt{google/} respectively. Qwen 2.5 and Llama 3.1 officially support context lengths up to 128k tokens. Gemma 3 supports up to 128k tokens but exhibited near-zero performance at this length across most configurations, so we evaluate Gemma 3 up to 64k only. We use 100 samples per configuration for Qwen and 50 for Llama and Gemma.}
    \label{tab:models}
\end{table}

\section{Computational Cost Analysis}
\label{sec:cost_breakdown}

We analyze computational cost using implementation-agnostic metrics that correlate with wall-clock time under optimized implementations: FLOPs for prefilling (compute-bound) and memory transfers for decoding (memory-bound). This approach avoids confounds from implementation-specific inefficiencies while capturing the fundamental cost structure.

\subsection{Cost Formulas}

For prefilling, which is compute-bound, we compute total FLOPs as:

\begin{align}
F_{\text{prefill}} &= B \cdot (F_{\text{emb}} + F_{\text{attn}} + F_{\text{mlp}} + F_{\text{logits}}) \\
F_{\text{emb}} &= 2 L d \\
F_{\text{attn}} &= N \big[2 L d (2d + 2 d_h n_{kv}) \notag \\
 &\qquad\; + \rho \, h L^2 (4 d_h + 3) \big] \\
F_{\text{mlp}} &= 2 N L d_{\text{mlp}} (3 d + 1) \\
F_{\text{logits}} &= 2 L d |V|
\end{align}

where $L$ is the sequence length, $d$ the hidden dimension, $h$ and $n_{kv}$ the number of query and KV heads, $d_h$ the head dimension, $N$ the number of layers, $d_{\text{mlp}}$ the MLP intermediate dimension, $|V|$ the vocabulary size, $B$ the batch size, and $\rho = 1-\text{sparsity}$ the attention density.

For decoding, which is memory-bound, we measure memory accesses:

\begin{align}
M_{\text{decode}} &= M_{\text{weights}} + B \cdot M_{\text{kv}} \\
M_{\text{kv}} &= 2 N L \, d_h \, n_{kv} \, \rho \\
M_{\text{weights}} &= N (4 d^2 + 3 d \, d_{\text{mlp}}) + d (|V|+1)
\end{align}

For sparse methods, we include importance-estimation overhead. Vertical-Slash uses
\begin{align}
F_{\text{VS}} &= B N h \, \big[2 d L q + 5 L q \notag \\
 &\qquad\; + 2 L \log_2 L + \tfrac{L}{64}(k_v + k_s) \big]
\end{align}
where $q$ is the number of queries used for importance estimation and $k_v$, $k_s$ are the numbers of selected vertical/slash patterns. Quest loads page representations
\begin{align}
M_{\text{Quest}} &= 2 B N \, n_{kv} \, d \cdot \tfrac{L}{p}
\end{align}
where $p$ is the page size (16 in our experiments).

\subsection{Prefilling: Sequence Length Drives Sparsity Impact}

\begin{figure*}[!t]
\centering
\includegraphics[width=\textwidth]{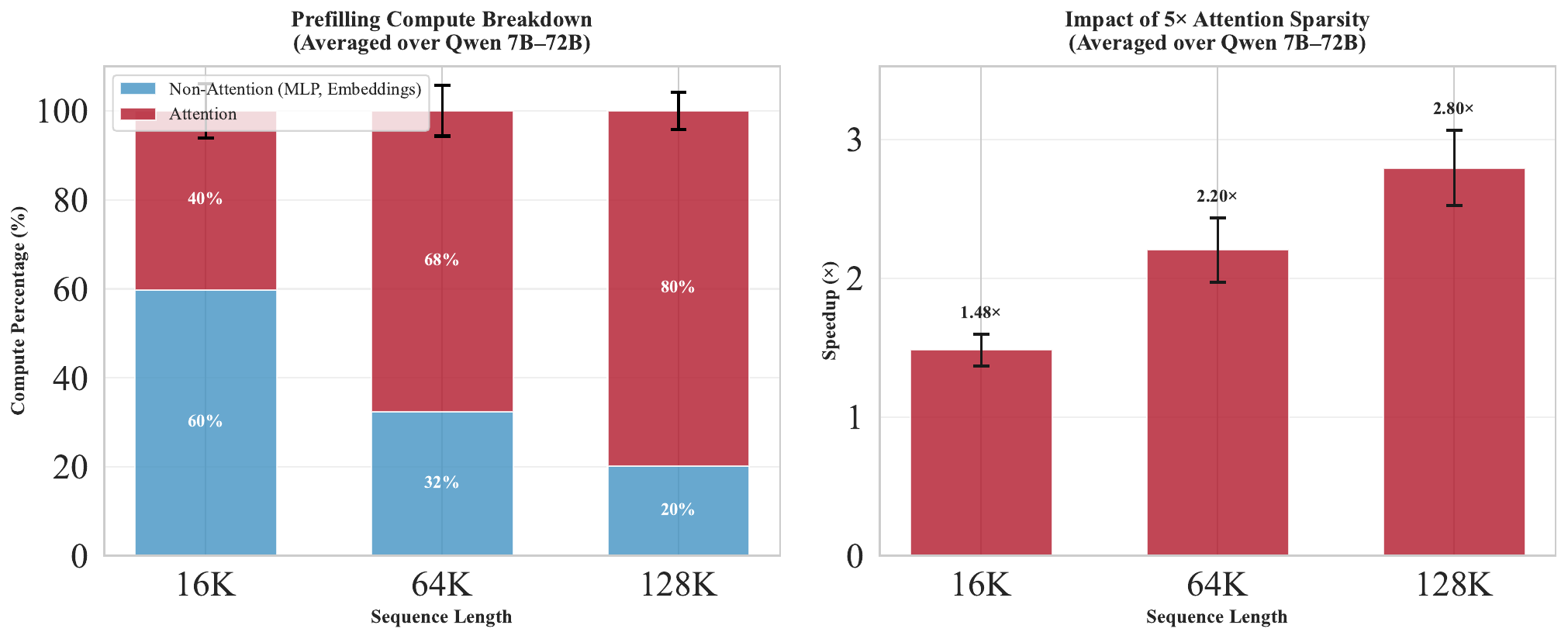}
\caption{Prefilling compute breakdown and sparsity benefits, averaged over Qwen 7B--72B with error bars showing standard deviation across model sizes. \textbf{Left}: As sequence length increases from 16K to 128K, attention grows from 40\% to 80\% of total FLOPs. \textbf{Right}: Consequently, 5$\times$ attention sparsity yields progressively greater speedups---from 1.5$\times$ at 16K to 2.8$\times$ at 128K.}
\label{fig:prefilling_seq_length_impact}
\end{figure*}

Attention cost scales quadratically with sequence length ($O(L^2)$) during prefilling, while non-attention costs (MLP, embeddings, logits) scale linearly ($O(L)$). This creates two regimes: at shorter sequences, non-attention components dominate; at longer sequences, attention becomes the primary cost.

Figure~\ref{fig:prefilling_seq_length_impact} illustrates the practical consequence. At 16K tokens, attention represents 40\% of prefilling FLOPs (averaged across Qwen 7B--72B), so 5$\times$ sparsity yields only 1.5$\times$ speedup. At 64K tokens, attention rises to 68\%, yielding 2.2$\times$ speedup. At 128K tokens, attention dominates at 80\%, enabling 2.8$\times$ speedup. Notably, the standard deviation across model sizes is small ($\pm$4--6\%), indicating this relationship holds regardless of model scale.

\subsection{Decoding: Sequence Length and Batch Size Both Matter}

\begin{figure*}[!t]
\centering
\includegraphics[width=\textwidth]{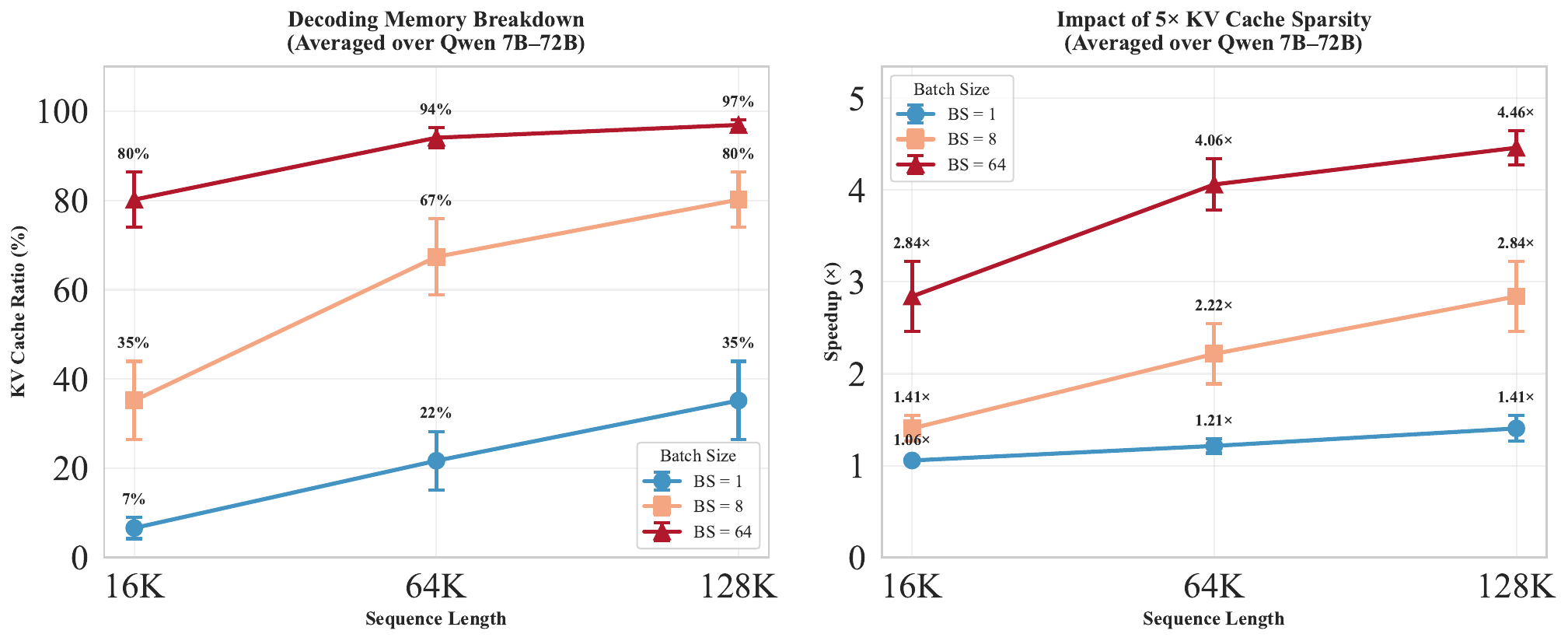}
\caption{Decoding cost breakdown and sparsity benefits, averaged over Qwen 7B--72B with error bars showing standard deviation across model sizes. \textbf{Left}: KV cache ratio increases with both sequence length and batch size. \textbf{Right}: Corresponding speedup from 5$\times$ KV cache sparsity. At batch size 1, sparse attention provides minimal benefit. At batch size 64, speedups reach 2.8--4.7$\times$.}
\label{fig:decoding_batch_size_dynamics}
\end{figure*}

Unlike prefilling, decoding cost depends on both sequence length and batch size. KV cache access scales linearly with context length ($O(L)$) and batch size ($O(B)$), while model weight loading is constant. This creates an important distinction: for prefilling, all cost components scale linearly with batch size, so the attention-to-total ratio remains constant. Decoding behaves differently---model weights are loaded once per forward pass regardless of batch size, while KV cache access scales with batch size.

Figure~\ref{fig:decoding_batch_size_dynamics} shows this clearly (averaged across Qwen 7B--72B). At batch size 1, weight loading dominates: KV cache represents only 7\% at 16K tokens, rising to 35\% at 128K. At batch size 8, the picture shifts: KV cache reaches 35--80\%. At batch size 64, KV cache dominates at 80--97\%, and sparse attention becomes highly effective with 2.8--4.7$\times$ speedups. The standard deviation across model sizes remains modest ($\pm$1--9\%), confirming these trends hold across model scales.

\subsection{Sliding-Window Architectures Need Longer Sequences}

\begin{figure}[!t]
\centering
\includegraphics[width=\linewidth]{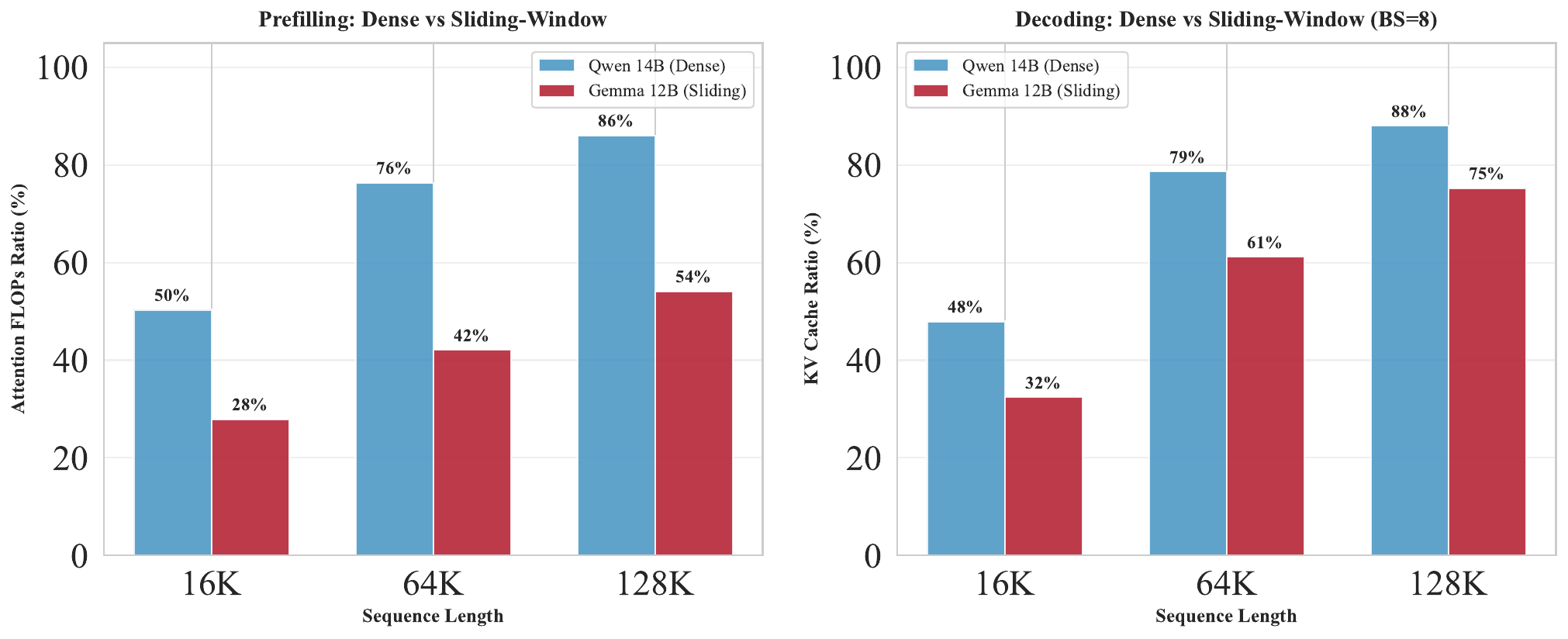}
\caption{Attention cost ratio comparison between Qwen 14B (dense attention) and Gemma 12B (sliding-window attention) at batch size 8. Gemma's architectural sparsity results in substantially lower attention ratios, requiring longer sequences for additional sparse attention to provide meaningful cost reduction.}
\label{fig:gemma_vs_qwen_comparison}
\end{figure}

Models with built-in architectural sparsity, such as Gemma 3's sliding-window attention (5 out of 6 layers use 1024-token windows), have lower baseline attention ratios. Figure~\ref{fig:gemma_vs_qwen_comparison} compares similar-sized models: at 64K tokens with batch size 8, Qwen 14B has 76\% attention ratio for prefilling versus Gemma 12B's 42\%. For decoding, Qwen reaches 79\% versus Gemma's 61\%. At 128K tokens, Gemma's attention ratio rises to 54\% for prefilling and 75\% for decoding.

This difference has practical implications for \isoCost comparisons. Because Gemma has a lower attention-to-total ratio at 64K, sparse prefilling reduces a smaller fraction of total FLOPs than for Qwen, so the dense--sparse frontiers overlap later (i.e., at higher costs / longer sequences). In principle, one would expect stronger overlap for Gemma at longer contexts as the attention ratio rises, but in our experiments most model--task configurations at 128K exhibit near-zero performance, preventing a meaningful prefilling comparison at that length. Decoding is less constrained: at sufficiently high batch size (e.g., $B=64$), KV-cache transfers dominate even for sliding-window models, so sparse decoding still yields practical benefits and can exhibit dense--sparse overlap in \isoCost space (see \cref{sec:results}).

\section{Comparison to Prior Work}
\label{app:prior_work}

Several concurrent works have explored evaluation of sparse attention methods. We summarise the key differences below.

\paragraph{SCBench \citep{Li2024SCBenchAK}} does not control for sequence length and evaluates at most two models from a single family, making it difficult to analyse the effects of sequence length and model size on sparse attention performance. Our work systematically varies sequence length (16K--128K tokens) and evaluates multiple model sizes within each of three model families.

\paragraph{\citet{liu2025can}} only considers models up to 10B parameters and does not address sparse attention in the prefilling phase. In contrast, we evaluate models up to 72B parameters and analyse both prefilling and decoding phases separately, revealing phase-specific behaviours.

\paragraph{\citet{Yuan2024KVCC}} tests sequence lengths only up to 32K tokens and includes models up to 10B parameters. Our evaluation extends to 128K tokens and 72B parameters, capturing the regime where sparse attention benefits are most pronounced.

In summary, our work is the first to systematically conduct an isoCost analysis for sparse attention, providing new insights into efficiency--accuracy trade-offs and generalisation across model size, sequence length, and sparsity.

\section{Extra Results}
\label{app:extra_results}

\subsection{Statistical Error Bounds}
\label{app:statistical_error}

We report standard error in \cref{fig:results_per_task} (main text, RQ2) for our per-task and per-method results. We omit standard error bars in isocost figures (\cref{fig:isocost_analysis}) for visual clarity, as the standard error is negligible. Here we derive the upper bound for the standard error across all configurations.

Since performance metrics lie in the $[0,1]$ range, the maximum standard deviation is $\sigma_{\max} = 0.5$ (achieved when the metric has a Bernoulli distribution with $p=0.5$). For configurations where we aggregate results over $N$ samples, the standard error is:

\begin{align}
SE = \frac{\sigma}{\sqrt{N}} \le \frac{\sigma_{\max}}{\sqrt{N}} = \frac{0.5}{\sqrt{N}}
\end{align}

In \cref{fig:isocost_analysis}, we aggregate performance across 9 tasks with 100 samples each (for Qwen), yielding $N=900$ samples total:

\begin{align}
SE_{\max} = \frac{0.5}{\sqrt{900}} = \frac{0.5}{30} \approx 0.0167
\end{align}

This upper bound of approximately $0.017$ is substantially smaller than the performance differences we observe between configurations (typically $>0.05$), justifying our decision to omit error bars for visual clarity in the isocost analysis.

\subsection{Per-Task Results by Model Family}
\label{app:per_task}

\begin{figure*}[p]
    \centering
    \includegraphics[width=\linewidth]{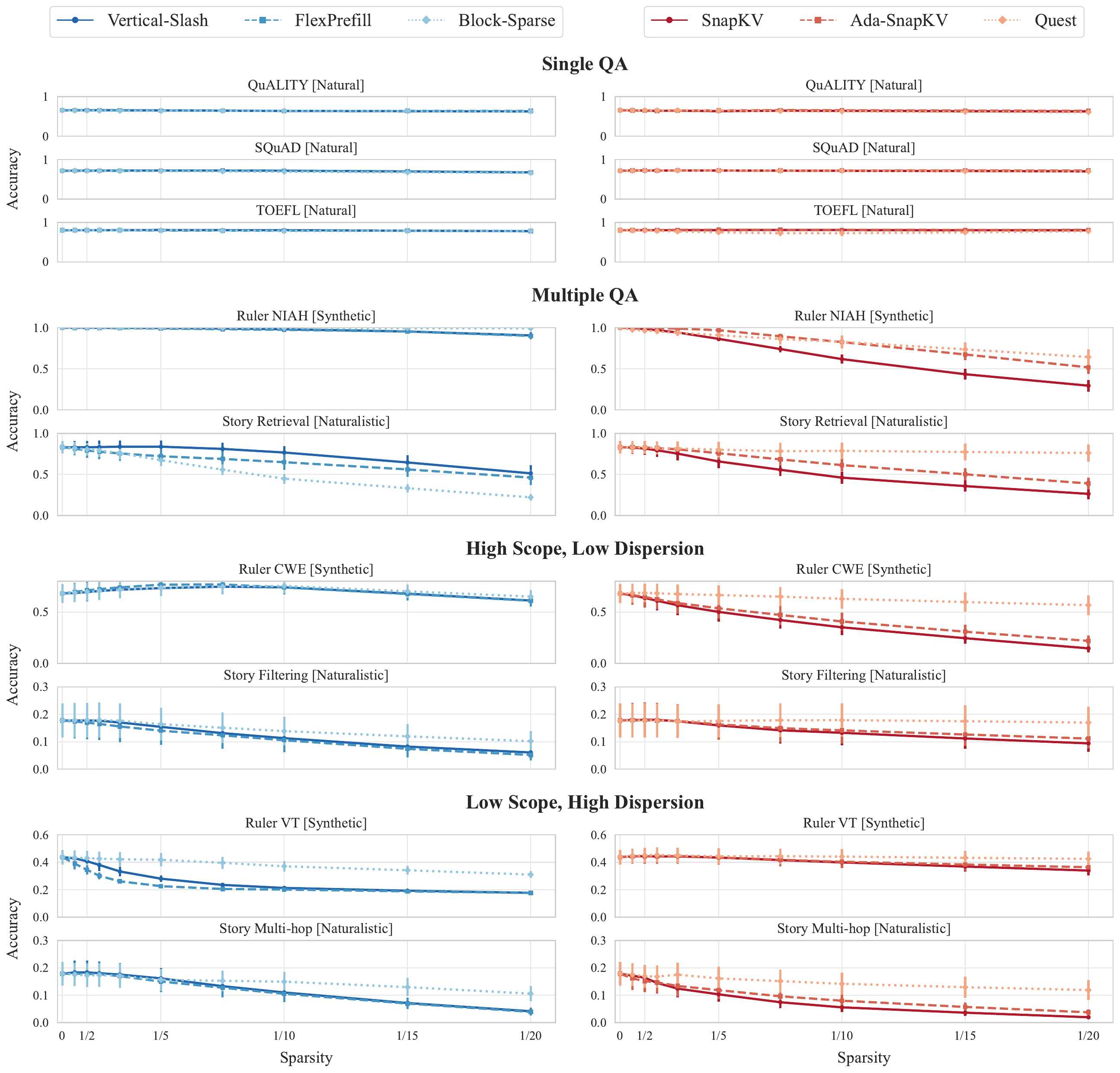}
    \caption{Per-task performance for \textbf{Qwen 2.5} models (7B, 14B, 32B, 72B) at sequence lengths 16k, 32k, and 64k. \textbf{Left}: prefilling methods. \textbf{Right}: decoding methods.}
    \label{fig:per_task_qwen}
\end{figure*}

\begin{figure*}[p]
    \centering
    \includegraphics[width=\linewidth]{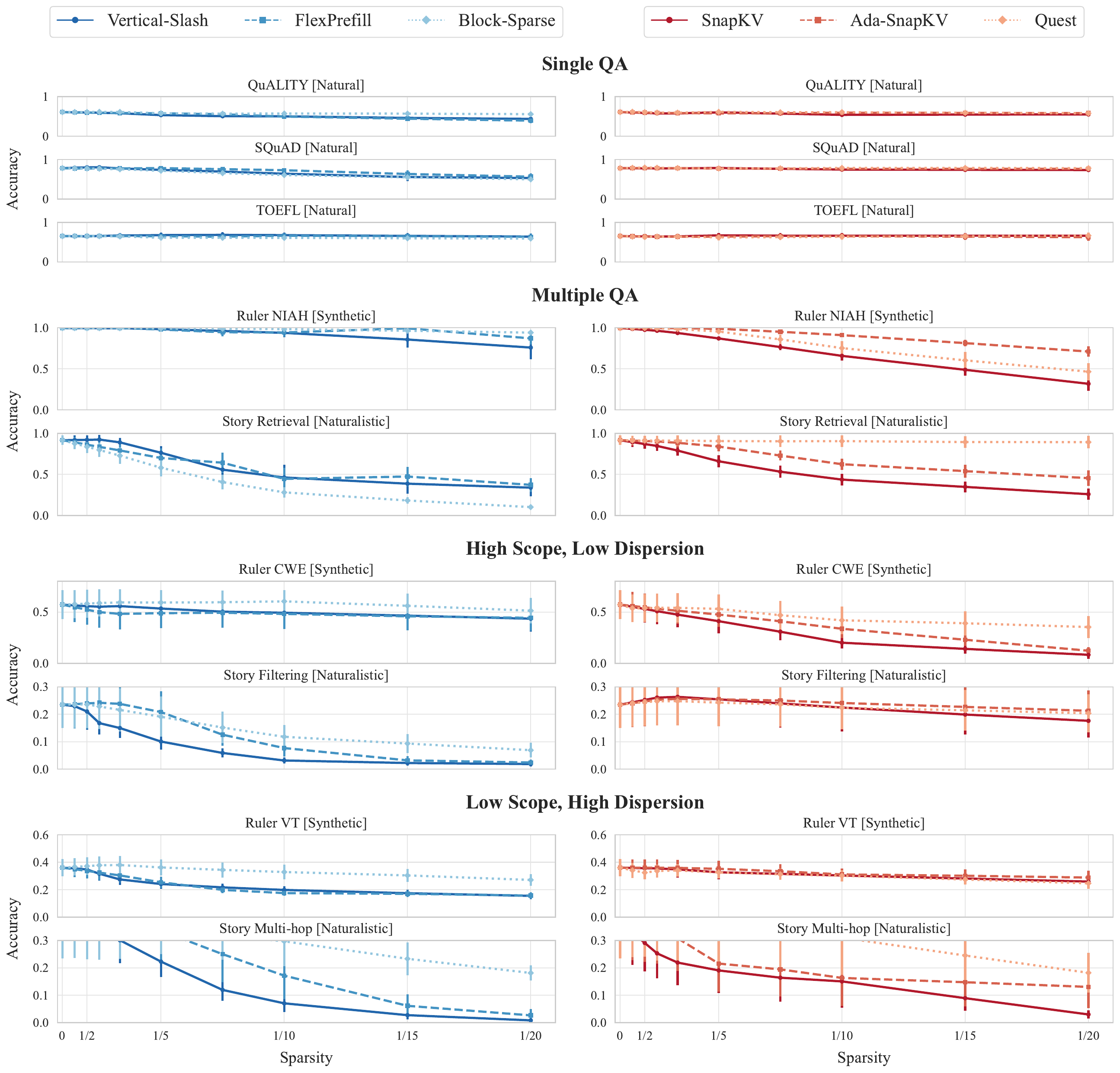}
    \caption{Per-task performance for \textbf{Llama 3.1} models (8B, 70B) at sequence lengths 16k, 32k, and 64k. \textbf{Left}: prefilling methods. \textbf{Right}: decoding methods.}
    \label{fig:per_task_llama}
\end{figure*}

\begin{figure*}[p]
    \centering
    \includegraphics[width=\linewidth]{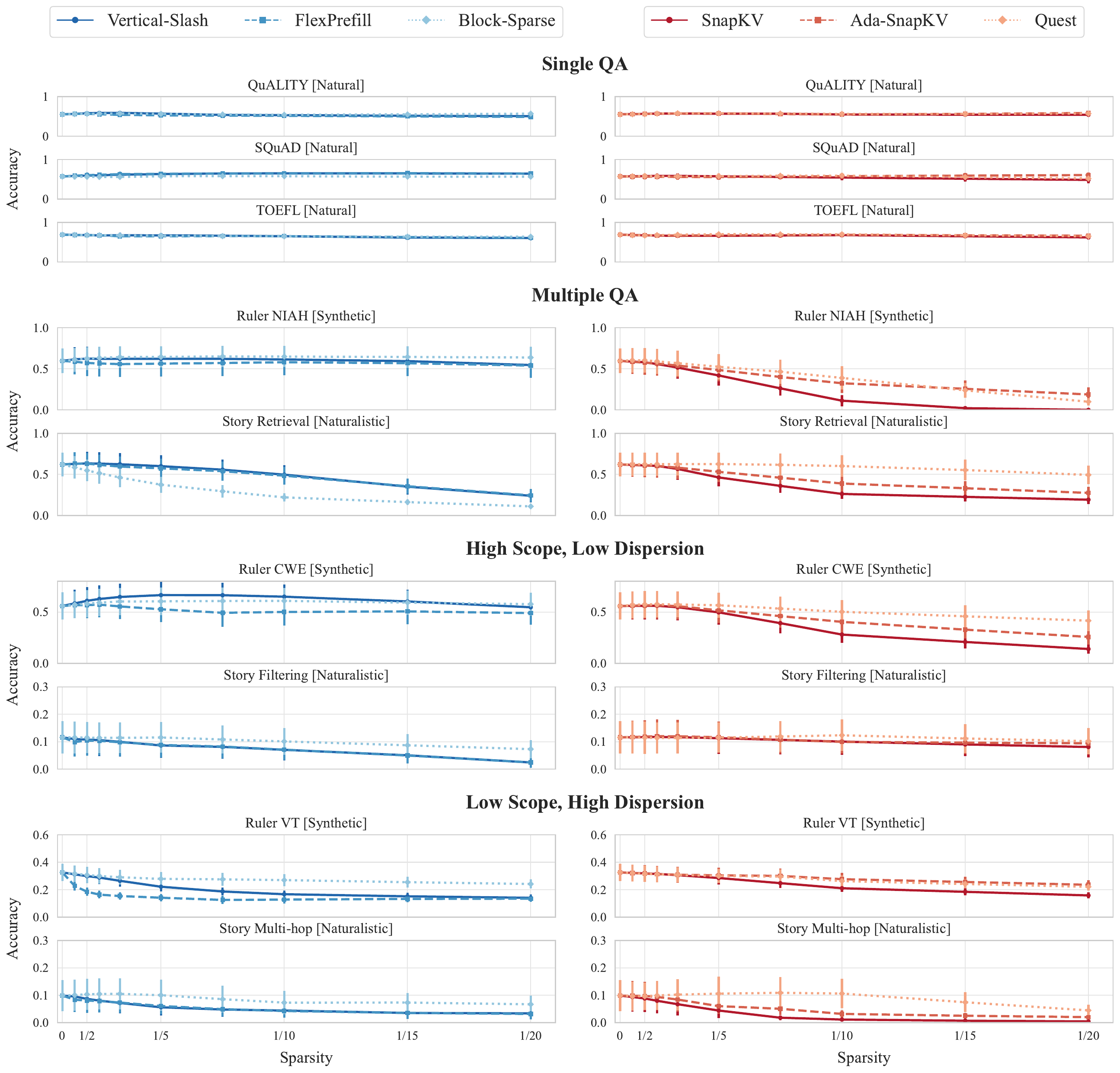}
    \caption{Per-task performance for \textbf{Gemma 3} models (4B, 12B, 27B) at sequence lengths 16k, 32k, and 64k. \textbf{Left}: prefilling methods. \textbf{Right}: decoding methods.}
    \label{fig:per_task_gemma}
\end{figure*}

This section provides per-task performance breakdowns for each model family, complementing the aggregated analysis in \cref{sec:results_individual}. \Cref{fig:per_task_qwen,fig:per_task_llama,fig:per_task_gemma} show results for Qwen 2.5, Llama 3.1, and Gemma 3 respectively.

\subsection{Sequence Length Effects}
\label{app:seq_length_absolute}

\begin{figure}[!t]
    \centering
    \includegraphics[width=\linewidth]{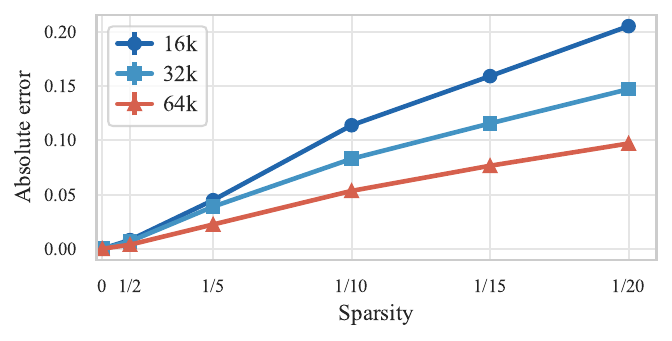}
    \caption{Absolute error vs.\ sparsity across sequence lengths. Results aggregated across all tasks, methods, and models (Qwen 2.5, Llama 3.1, Gemma 3).}
    \label{fig:seq_length_absolute}
\end{figure}

\Cref{fig:seq_length_absolute} presents the absolute error perspective on sequence length effects, complementing the relative error analysis in \cref{sec:results_seq_length}. The absolute error is $\bar{p}_{\text{dense}} - \bar{p}_{\text{sparse}}$, where $\bar{p}$ denotes mean performance. The pattern mirrors the relative error findings: longer sequences tolerate higher sparsity with smaller absolute performance degradation.

\begin{figure*}[t!]
    \centering
    \includegraphics[width=\linewidth]{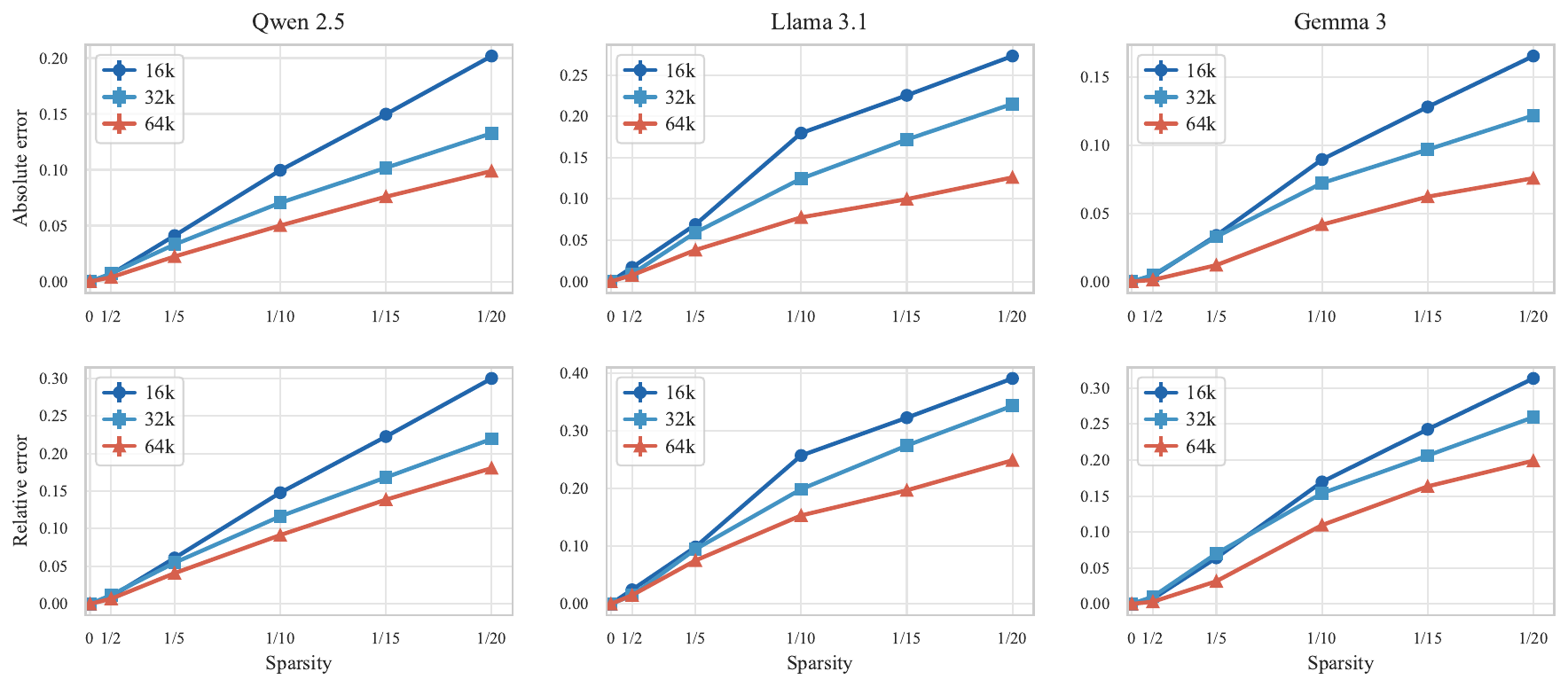}
    \caption{Sequence length effects on sparsity tolerance by model family. \textbf{Top row}: absolute error vs.\ sparsity. \textbf{Bottom row}: relative error vs.\ sparsity. Results aggregated across all tasks and methods within each family.}
    \label{fig:seq_length_per_family}
\end{figure*}

\Cref{fig:seq_length_per_family} provides per-family breakdowns of the sequence length analysis. The trend of improved sparsity tolerance at longer sequences holds consistently across all three model families, with minor variations in magnitude.

\subsection{Model Size Analysis}
\label{app:model_size_analysis}

\begin{figure*}[t!]
    \centering
    \includegraphics[width=\linewidth]{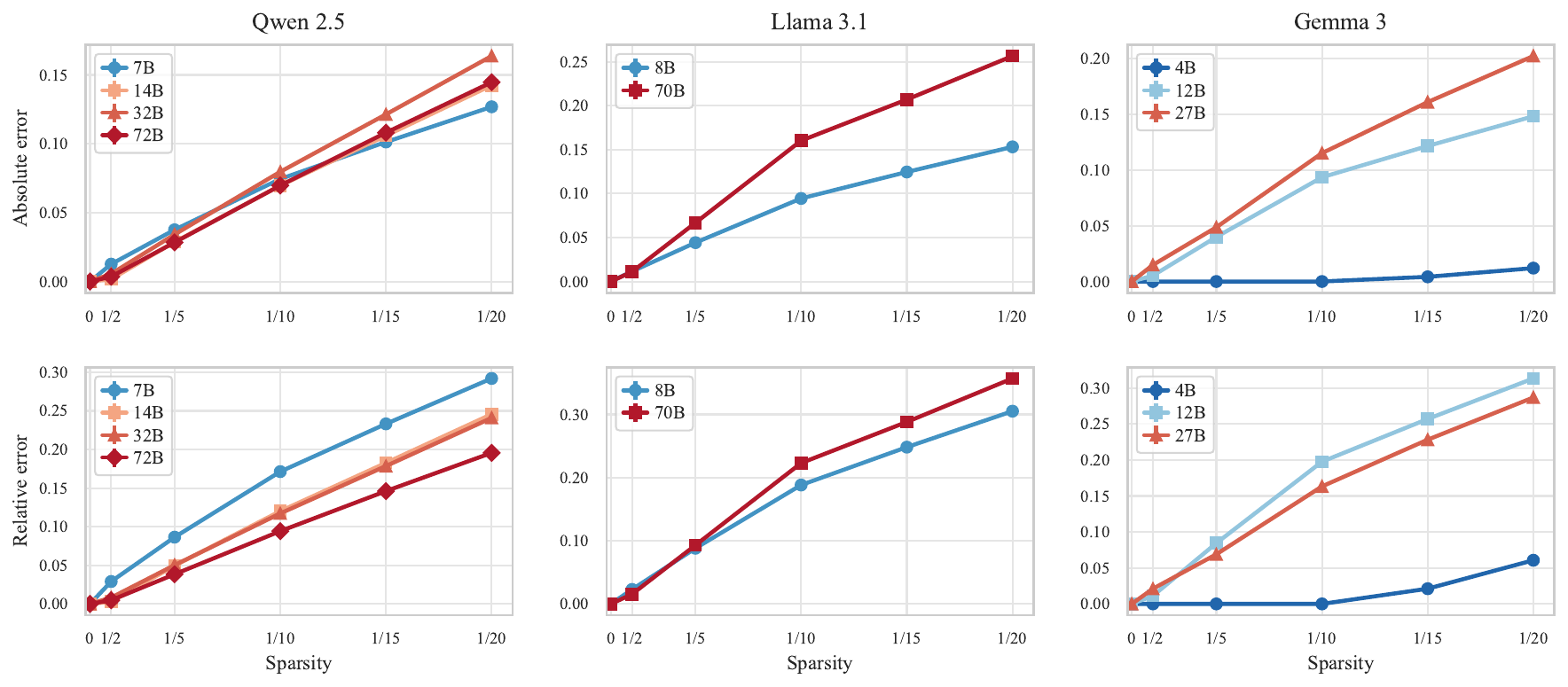}
    \caption{Model size effects on sparsity tolerance aggregated across all tasks. \textbf{Top row}: absolute error vs.\ sparsity. \textbf{Bottom row}: relative error vs.\ sparsity. Results aggregated across all tasks, methods, and sequence lengths 16--64k for each model family.}
    \label{fig:model_size_aggregate}
\end{figure*}

We analyse how sparsity tolerance varies with model scale. \Cref{fig:model_size_aggregate} shows model size effects aggregated across all tasks, methods, and sequence lengths for each model family. On average, model size shows no clear correlation with sparsity tolerance---the lines for different model sizes largely overlap, indicating that larger models do not systematically tolerate more or less sparsity than smaller ones.

\begin{figure*}[t!]
    \centering
    \includegraphics[width=\linewidth]{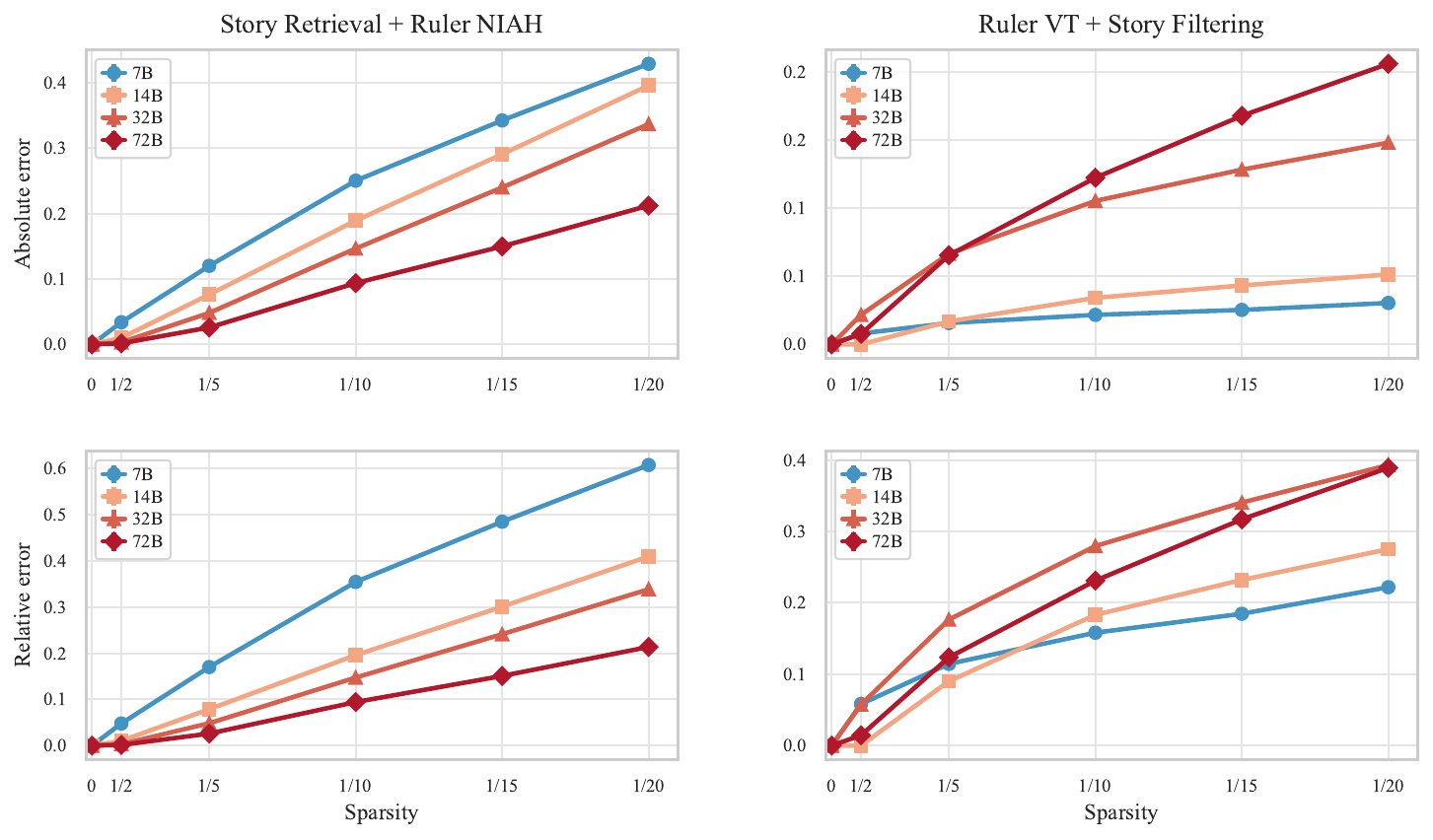}
    \caption{Model size effects on sparsity tolerance for Qwen 2.5 (7B--72B). Absolute error is $\bar{p}_{\text{dense}} - \bar{p}_{\text{sparse}}$; relative error is $(\bar{p}_{\text{dense}} - \bar{p}_{\text{sparse}}) / \bar{p}_{\text{dense}}$, where $\bar{p}$ denotes mean performance. \textbf{Top row}: absolute error vs.\ sparsity. \textbf{Bottom row}: relative error vs.\ sparsity. \textbf{Left column}: easy tasks (Story Retrieval, Ruler NIAH). \textbf{Right column}: hard tasks (Ruler VT, Story Filtering). Results aggregated across methods and sequence lengths 16--64k.}
    \label{fig:model_size_analysis}
\end{figure*}

However, this aggregate finding masks important task-dependent patterns revealed in \cref{fig:model_size_analysis}.

\paragraph{Model size effects depend on task difficulty.} \Cref{fig:model_size_analysis} presents contrasting perspectives on model size effects. On tasks where all model sizes achieve near-perfect dense accuracy (left column: Story Retrieval, Ruler NIAH), larger models tolerate more sparsity---at sparsity 0.95 ($1/20$ budget), 72B shows 0.20 absolute error compared to 0.50 for 7B. Conversely, on challenging tasks where dense accuracy scales with model size (right column: Ruler VT, Story Filtering), larger models exhibit \textit{larger} absolute errors at equivalent sparsity---72B shows 0.21 absolute error while 7B shows only 0.03 at sparsity 0.95. The relative error perspective (bottom row) shows consistent patterns: larger models have lower relative error on easy tasks but higher relative error on hard tasks.

These divergent patterns arise from how sparsity interacts with model capacity. Sparse attention reduces effective model capacity by limiting information flow. When a model operates far above a task's difficulty threshold, this capacity reduction has minimal impact on outputs. When model capacity approximately matches task difficulty, even modest sparsity degrades performance. Larger models achieve higher dense accuracy on difficult tasks, operating closer to their capacity limits on these tasks---making them more vulnerable to capacity reductions from sparsity. Evaluations on tasks where models achieve perfect or near-perfect accuracy---common in benchmarks like Needle-in-a-Haystack---cannot reveal these vulnerabilities.

\section{Use of AI Assistants}
\label{sec:ai_assistants}
We used Claude Opus 4.5 for grammar and style suggestions during the writing of this paper. All scientific content, analysis, and conclusions are the authors' own work.

\clearpage
\onecolumn
\section{Prompt Template}
\label{sec:prompt_template}

\begin{tcolorbox}[colback=gray!10, colframe=black, sharp corners=south, boxrule=0.5mm]
\scriptsize
\textbf{Input format:}
\begin{verbatim}
You are provided with a task introduction, context, and a question.

{task_intro}

Below is your question. I will state it both before and after the context.

<question>
{question}
</question>

<context>
{context}
</context>

<question_repeated>
{question}
</question_repeated>

Instructions:
1. First, provide a brief explanation of your reasoning process. Explain how you identified
   the relevant information from the context and how you determined your answer.
2. Then, provide your final answer following this exact format:
<answer>
{answer_format}
</answer>

Your response must follow this structure exactly:
<explanation>
Your explanation here...
</explanation>
<answer>
Your answer here...
</answer>

Important:
{extra_instructions}
- Keep your explanations clear, coherent, concise, and to the point.
- Do not include any additional text, explanations, or reasoning in the answer section.
\end{verbatim}
\end{tcolorbox}

\clearpage
\section{Example Story Narrative}
\label{sec:example_story_narrative}

\begin{tcolorbox}[colback=gray!10, colframe=black, fonttitle=\bfseries, sharp corners=south, boxrule=0.5mm]
\scriptsize
\begin{verbatim}
\end{verbatim}
Chapter 1:
\\
Beneath gentle breezes, Arion ventured into Athens, curious about its secrets. Long journeys had led Arion to Athens, a step closer to understanding. Soon enough, a tense negotiation seized everyone's attention. Cleo appeared as if expecting Arion, engaging them without delay. Carefully, they navigated the topic of old feuds, wary of awakening dormant animosities that still simmered. In a calm moment, they compared notes on the traders who passed through Athens, each leaving their subtle mark. In hushed tones, they spoke of local customs and distant rumors, sharing hints of hidden pathways. Following subtle bargaining with Cleo, Arion claimed ownership of lavish crystal lamp. With a light gesture, Arion acknowledged Cleo once more before departing. Nothing would be the same as Arion left Athens, thoughts turning inward. In quiet corners, ambitions simmered, waiting for a spark. 
\\
\\
Chapter 2:
\\
At dawn, Arion reached the gates of Hippo Regius, where merchants and travelers converged. This place might hold a clue Arion had long sought. Hardly had Arion arrived before a violent storm stirred uneasy whispers. Thanos approached Arion, eyes bright with opportunity. They lingered over tales of old alliances and forgotten disputes, weaving past into present. They debated the meaning of recent events, each seeking patterns in the chaos. Their reflections turned to the interplay of supply and demand, seeing how fortunes might turn in an instant. After reaching terms with Thanos, Arion took possession of ceremonial gold seal. Arion turned from Thanos, ready to move on. In parting, Arion acknowledged that the journey still had far to run. Hidden corners of the city promised knowledge or peril. 
\\
\\
Chapter 3:
\\
The threshold of Emerita Augusta welcomed Arion, who felt the weight of untold stories. Arion came here hoping to learn something new, or perhaps gain an advantage. Within hours, a violent storm disrupted the familiar routines. There, Arion encountered Niko, who seemed eager to exchange words or goods. Their words lingered on rumors of distant lands, where fortunes or ruin awaited bold seekers. They debated the meaning of recent events, each seeking patterns in the chaos. Their dialogue danced around subtle clues, each suggestion hinting at treasures undiscovered. The transaction concluded with Arion acquiring delicate porcelain sword from Niko. With a light gesture, Arion acknowledged Niko once more before departing. Eventually, Arion moved on, carrying new impressions forward. The distant hum of voices hinted at unseen deals. 
\\
\\
Chapter 4:
\\
Under fading daylight, Arion set foot in Berenice, eager to learn what it offered. A quiet determination brought Arion to Berenice, ever searching for meaning. a sudden market crash cast its shadow over Berenice, changing plans and minds. Roxana approached Arion, eyes bright with opportunity. Together, they reflected on the nature of trust and deceit, aware that fate often twists. They compared accounts of strange visitors bearing knowledge or confusion, each arrival a new riddle in Berenice. A short exchange revealed uncharted corners of Berenice, where knowledge or secrets might dwell. mystic bronze lamp changed hands as Arion completed the purchase from Roxana. Arion handed over lavish crystal lamp to Roxana as the deal closed. With a light gesture, Arion acknowledged Roxana once more before departing. As Arion prepared to depart, the path ahead remained uncertain but compelling. Somewhere, a whisper promised answers for those who dared. 
\\
\\
Chapter 5:
\\
Under fading daylight, Arion set foot in Syracuse, eager to learn what it offered. In pursuit of truth, Arion looked to Syracuse for subtle revelations. Not long after arriving, an opulent banquet shook the local order. Phaedra appeared as if expecting Arion, engaging them without delay. Their words traced over delicate negotiations that had once sealed lasting truces in Syracuse. Carefully, they navigated the topic of old feuds, wary of awakening dormant animosities that still simmered. They delved into the subtle art of earning trust in a place where trust was scarce and hard-won. With measured consideration, Arion purchased engraved emerald goblet from Phaedra, examining it closely. In quiet understanding, Arion left Phaedra, their paths diverging. In parting, Arion acknowledged that the journey still had far to run. A subtle tension lingered, as though fate held its breath. 
\end{tcolorbox}

\clearpage
\section{Example Task Inputs}
\label{sec:example_task_inputs}

\subsection{Question Answering (QA)}
\label{example_task_input:qa}

\begin{tcolorbox}[colback=gray!10, colframe=black, fonttitle=\bfseries, sharp corners=south, before skip=0pt, after skip=0pt, boxrule=0.5mm]
\tiny
\textbf{Input format:}
\begin{verbatim}
    I will provide you with multiple documents and ask you a question about one specific document.
    
    Below is your question. I will state it both before and after the context.
    
    <question>
    Question about document 39:
    Who works to get workers higher compensation?
    </question>
    
    <context>
    Document 1:
    [...text omitted...]
    
    Document 39:
    Jobs with high demand and low supply pay more. Professional and labor organizations
    can raise wages by limiting worker supply and using collective bargaining or political influence.
    
    Document 47:
    [...text omitted...]
    </context>
    
    <question_repeated>
    Question about document 39:
    Who works to get workers higher compensation?
    </question_repeated>
    
    Instructions:
    1. Provide a brief explanation of your reasoning process.
    2. Then, give your final answer in this format:
    <answer>
    Your answer here...
    </answer>
    
    Your response must follow this structure:
    <explanation>
    Your explanation here...
    </explanation>
    <answer>
    Your answer here...
    </answer>
    
    Important:
    - Do not use complete sentences in the answer.
    - For dates: Include ONLY the COMPLETE date if specifically asked.
    - For locations: Use the shortest unambiguous form (e.g., 'New York' not 'New York City').
    - For comparisons: State ONLY the answer that matches the criteria
    - Keep your explanations clear, coherent, concise, and to the point.
    - Do not include any additional text, explanations, or reasoning in the answer section.
\end{verbatim}

\vspace{0.5em}

\textbf{Example answer:}
\begin{verbatim}
    <explanation>
    I found the relevant sentence in document 39, which states that professional and labor
    organizations help increase wages using bargaining and political means.
    </explanation>
    <answer>
    Professional and labor organizations
    </answer>
\end{verbatim}
\end{tcolorbox}

\newpage
\subsection{RULER - Needle-in-a-Haystack (NIAH)}
\label{example_task_input:niah}

\begin{tcolorbox}[colback=gray!10, colframe=black, fonttitle=\bfseries, sharp corners=south, boxrule=0.5mm]
    \tiny
    \textbf{Input format:}
    \begin{verbatim}
    I will provide you with a document containing multiple key-value pairs. 
    Your task is to extract specific values associated with given keys.

    Below are your questions. I will state them both before and after the context.

    <questions>
    Extract the values for the following keys:
    key-A, key-B, key-C, key-D
    </questions>

    <context>
    The value for key-A is: value-A.
    The value for key-X is: value-X.
    The value for key-B is: value-B.
    The value for key-Y is: value-Y.
    The value for key-C is: value-C.
    The value for key-Z is: value-Z.
    The value for key-D is: value-D.
    </context>

    <questions_repeated>
    Extract the values for the following keys:
    key-A, key-B, key-C, key-D
    </questions_repeated>

    Instructions:
    1. First, provide a brief explanation of your reasoning process. Explain how you identified 
       the relevant information from the context and how you determined your answer.
    2. Then, provide your final answer following this exact format:
    <answer>
    1. The answer for <key1> is <value1>.
    2. The answer for <key2> is <value2>.
    etc.
    </answer>

    Your response must follow this structure exactly:
    <explanation>
    Your explanation here...
    </explanation>
    <answer>
    Your answer here...
    </answer>

    Important:
    - Provide answers in the exact order of the requested keys
    - Each answer must follow the format: "<number>. The answer for <key> is <value>."
    - Ensure exact key matches - do not modify or paraphrase the keys
    - Values must match exactly as they appear in the document
    - Keep your explanations clear, coherent, concise, and to the point.
    - Do not include any additional text, explanations, or reasoning in the answer section.
    \end{verbatim}
    
    \vspace{0.5em}
    
    \textbf{Example answer:}
    \begin{verbatim}
    <explanation>
    I scanned the context for exact matches of the requested keys. For each key, I extracted 
    the value as stated directly after the pattern "The value for key-X is: ...".
    </explanation>
    <answer>
    1. The answer for key-A is value-A.
    2. The answer for key-B is value-B.
    3. The answer for key-C is value-C.
    4. The answer for key-D is value-D.
    </answer>
    \end{verbatim}
\end{tcolorbox}

\newpage
\subsection{RULER - Common Word Extraction (CWE)}
\label{example_task_input:cwe}

\begin{tcolorbox}[colback=gray!10, colframe=black, fonttitle=\bfseries, sharp corners=south, boxrule=0.5mm]
\tiny
\textbf{Input format:}
\begin{verbatim}
    You will be given a numbered list of words. Your task is to identify the most 
    frequently occurring words. You should solve this task by carefully reading 
    and analyzing the word list. Do not attempt to write code or use programming 
    tools to count frequencies. This is a test of your ability to track word 
    frequencies directly.
    
    Below is your question. I will state it both before and after the context.
    
    <question>
    The list contains exactly 10 words that appear 30 times each.
    All other words appear 3 times each.
    Your task is to identify the 10 words that appear 30 times each.
    </question>
    
    <context>
    1. alpha
    2. beta
    3. gamma
    4. delta
    5. alpha
    6. epsilon
    ...
    [...list continues with randomized repeated words...]
    ...
    N. gamma
    </context>
    
    <question_repeated>
    The list contains exactly 10 words that appear 30 times each.
    All other words appear 3 times each.
    Your task is to identify the 10 words that appear 30 times each.
    </question_repeated>
    
    Instructions:
    1. First, provide a brief explanation of your reasoning process.
       Explain how you identified the relevant information from the context 
       and how you determined your answer.
    2. Then, provide your final answer following this exact format:
    <answer>
    1. word_one
    2. word_two
    ...
    10. word_ten
    </answer>
    
    Your response must follow this structure exactly:
    <explanation>
    Your explanation here...
    </explanation>
    <answer>
    Your answer here...
    </answer>
    
    Important:
    - List exactly 10 words, one per line, numbered from 1 to 10.
    - Keep your explanations clear, coherent, concise, and to the point.
    - Do not include any additional text, explanations, or reasoning in the answer section.
\end{verbatim}

\vspace{0.5em}

\textbf{Example answer:}
\begin{verbatim}
    <explanation>
    I scanned the word list and tracked the frequency of each word. 
    The following 10 words appeared 30 times each, which I confirmed by careful counting.
    </explanation>
    <answer>
    1. diligent
    2. ash
    3. pour
    4. chateau
    5. marble
    6. laparoscope
    7. grub
    8. vinyl
    9. mobility
    10. kettledrum
    </answer>
\end{verbatim}
\end{tcolorbox}

\newpage
\subsection{RULER - Variable Tracking (VT)}
\label{example_task_input:vt}

\begin{tcolorbox}[colback=gray!10, colframe=black, fonttitle=\bfseries, sharp corners=south, boxrule=0.5mm]
\tiny
\textbf{Input format:}
\begin{verbatim}
    I will provide you with a text containing variable assignments. The text contains two types of assignments:
    1. Numeric assignments that set a variable to a number (e.g., "VAR ABC = 12345")
    2. Copy assignments that set a variable equal to another variable (e.g., "VAR XYZ = VAR ABC")
    Variables are sequences of uppercase letters. The assignments can appear in any order in the text.
    
    Below is your question. I will state it both before and after the context.
    
    <question>
    Which variables resolve to the value 41015? A variable resolves to 41015 if it is either directly assigned
    41015, or assigned to another variable that resolves to 41015.
    </question>
    
    <context>
    VAR A = VAR B
    VAR B = 41015
    VAR C = VAR D
    VAR D = VAR B
    VAR E = 12345
    VAR F = VAR G
    VAR G = VAR H
    VAR H = VAR B
    </context>
    
    <question_repeated>
    Which variables resolve to the value 41015? A variable resolves to 41015 if it is either directly assigned
    41015, or assigned to another variable that resolves to 41015.
    </question_repeated>
    
    Instructions:
    1. First, provide a brief explanation of your reasoning process. Explain how you identified 
       the relevant information from the context and how you determined your answer.
    2. Then, provide your final answer following this exact format:
    <answer>
    VARIABLE_ONE VARIABLE_TWO etc.
    </answer>
    
    Your response must follow this structure exactly:
    <explanation>
    Your explanation here...
    </explanation>
    <answer>
    Your answer here...
    </answer>
    
    Important:
    - List ONLY the variable names that resolve to the target value.
    - Variables can be listed in any order.
    - Do not include "VAR" prefix in your answer. Do not include punctuation.
    - Keep your explanations clear, coherent, concise, and to the point.
    - Do not include any additional text, explanations, or reasoning in the answer section.
\end{verbatim}

\vspace{0.5em}

\textbf{Example answer:}
\begin{verbatim}
    <explanation>
    I traced each variable assignment to see if it leads to the value 41015. B is directly assigned 41015.
    A, D, and H point to B. C and G point to D and H, respectively. So A B C D G H resolve to 41015.
    </explanation>
    <answer>
    A B C D G H
    </answer>
\end{verbatim}
\end{tcolorbox}

\newpage
\subsection{Story Retrieval}
\label{example_task_input:story_retrieval}

\begin{tcolorbox}[colback=gray!10, colframe=black, fonttitle=\bfseries, sharp corners=south, boxrule=0.5mm]
\tiny
\textbf{Input format:}
\begin{verbatim}
    You are given a narrative composed of multiple chapters. Throughout these chapters, the 
    protagonist travels between different locations, meets various characters, and engages 
    in trading activities. All items mentioned in the narrative are unique, and their 
    ownership can change through trades. Your task is to carefully read the narrative and 
    answer the questions based on the provided information.
    
    Below are your questions. I will state them both before and after the context.
    
    <questions>
    1. In Chapter 3, which character did the protagonist interact with?
    2. In Chapter 5, which specific item was acquired by the protagonist?
    3. In Chapter 7, which specific location did the protagonist visit?
    </questions>
    
    <context>
    Chapter 1:
    [...text omitted...]
    
    Chapter 3:
    Arion entered Babylon and met Thanos. After exchanging stories, Arion acquired a silver idol.
    
    Chapter 5:
    In Berenice Troglodytica, Arion encountered Xanthe and traded for a golden vase.
    
    Chapter 7:
    Delphi welcomed Arion with quiet mystery. A meeting with Vitalis ended with a jade idol.
    </context>
    
    <questions_repeated>
    1. In Chapter 3, which character did the protagonist interact with?
    2. In Chapter 5, which specific item was acquired by the protagonist?
    3. In Chapter 7, which specific location did the protagonist visit?
    </questions_repeated>
    
    Instructions:
    1. First, provide a brief explanation of your reasoning process. Explain how you identified
       the relevant information from the context and how you determined your answer.
    2. Then, provide your final answer following this exact format:
    <answer>
    1. ANSWER_ONE
    2. ANSWER_TWO
    etc.
    </answer>
    
    Your response must follow this structure exactly:
    <explanation>
    Your explanation here...
    </explanation>
    <answer>
    Your answer here...
    </answer>
    
    Important:
    - For answers, use one line per answer with the number prefix
    - Do not include articles like 'the' or 'a' in answers
    - Answers should be specific names/items/locations mentioned in the text
    - Keep your explanations clear, coherent, concise, and to the point.
    - Do not include any additional text, explanations, or reasoning in the answer section.
\end{verbatim}

\vspace{0.5em}

\textbf{Example answer:}
\begin{verbatim}
    <explanation>
    I located Chapter 3 in the context and identified Thanos as the mentioned character.
    In Chapter 5, Arion acquired a golden vase from Xanthe.
    Chapter 7 stated that Arion visited Delphi, so I used that as the answer.
    </explanation>
    <answer>
    1. Thanos
    2. Golden Vase
    3. Delphi
    </answer>
\end{verbatim}
\end{tcolorbox}

\newpage
\subsection{Story Filtering}
\label{example_task_input:story_filtering}

\begin{tcolorbox}[colback=gray!10, colframe=black, fonttitle=\bfseries, sharp corners=south, boxrule=0.5mm]
\tiny
\textbf{Input format:}
\begin{verbatim}
    You are given a narrative composed of multiple chapters. Throughout these chapters, the 
    protagonist travels between different locations, meets various characters, and engages 
    in trading activities. All items mentioned in the narrative are unique, and their 
    ownership can change through trades. Your task is to carefully read the narrative and 
    answer the questions based on the provided information.
    
    Below is your question. I will state it both before and after the context.
    
    <question>
    Identify all chapters where the protagonist did not buy any item.
    Note: There are exactly 2 chapters without any purchases.
    </question>
    
    <context>
    Chapter 1:
    [... Arion visits Athens and purchases a crystal lamp ...]
    
    Chapter 2:
    [... Arion travels to Hippo Regius and buys a gold seal ...]
    
    Chapter 3:
    [... Arion enters Babylon and engages in an ongoing event but do not buy anything ...]
    
    Chapter 4:
    [... Arion arrives in Pergamon and has conversations, but no purchases are mentioned ...]
    
    Chapter 5:
    [... Arion goes to Delphi and buys a jade idol ...]
    </context>
    
    <question_repeated>
    Identify all chapters where the protagonist did not buy any item.
    Note: There are exactly 2 chapters without any purchases.
    </question_repeated>
    
    Instructions:
    1. First, provide a brief explanation of your reasoning process. Explain how you identified 
       the relevant information from the context and how you determined your answer.
    2. Then, provide your final answer following this exact format:
    <answer>
    chapter_id_1, chapter_id_2, ...
    </answer>
    
    Your response must follow this structure exactly:
    <explanation>
    Your explanation here...
    </explanation>
    <answer>
    Your answer here...
    </answer>
    
    Important:
    - In the answer section, provide only the chapter IDs separated by commas.
    - Keep your explanations clear, coherent, concise, and to the point.
    - Do not include any additional text, explanations, or reasoning in the answer section.
\end{verbatim}

\vspace{0.5em}

\textbf{Example answer:}
\begin{verbatim}
    <explanation>
    I scanned each chapter to check whether a purchase by the protagonist was explicitly 
    described. In Chapter 3 and 4, no item acquisition are mentioned. Other chapters include 
    phrases like "Arion purchased" or "Arion acquired", indicating a transaction.
    </explanation>
    <answer>
    3, 4
    </answer>
\end{verbatim}
\end{tcolorbox}

\newpage
\subsection{Story Multi-hop}
\label{example_task_input:story_multihop}

\begin{tcolorbox}[colback=gray!10, colframe=black, fonttitle=\bfseries, sharp corners=south, boxrule=0.5mm]
\tiny
\textbf{Input format:}
\begin{verbatim}
    You are given a narrative composed of multiple chapters. Throughout these chapters,
    the protagonist travels between different locations, meets various characters, 
    and engages in trading activities. All items mentioned in the narrative are unique, 
    and their ownership can change through trades. Your task is to carefully read the 
    narrative and answer the questions based on the provided information.
    
    Below is your question. I will state it both before and after the context.
    
    <question>
    What was the last item that the protagonist acquired before acquiring timeworn amber sword?
    </question>
    
    <context>
    Chapter 1:
    [... narrative text omitted for brevity ...]
    
    Chapter 17:
    The transaction concluded with Arion acquiring pristine bronze seal from Damon.
    
    Chapter 18:
    After reaching terms with Marcus, Arion took possession of timeworn amber sword.
    </context>
    
    <question_repeated>
    What was the last item that the protagonist acquired before acquiring timeworn amber sword?
    </question_repeated>
    
    Instructions:
    1. First, provide a brief explanation of your reasoning process. Explain how you identified 
       the relevant information from the context and how you determined your answer.
    2. Then, provide your final answer following this exact format:
    <answer>
    ITEM_NAME
    </answer>
    
    Your response must follow this structure exactly:
    <explanation>
    Your explanation here...
    </explanation>
    <answer>
    Your answer here...
    </answer>
    
    Important:
    - Provide only the item name in the answer section.
    - Do not include articles like 'the' or 'a' in your answer.
    - The item name must be exactly as mentioned in the text.
    - Keep your explanations clear, coherent, concise, and to the point.
    - Do not include any additional text, explanations, or reasoning in the answer section.
\end{verbatim}

\vspace{0.5em}

\textbf{Example answer:}
\begin{verbatim}
    <explanation>
    I located the chapter where the protagonist acquired the timeworn amber sword.
    Then, I scanned earlier chapters to find the most recent prior acquisition, 
    which occurred in Chapter 17 with the item pristine bronze seal.
    </explanation>
    <answer>
    pristine bronze seal
    </answer>
\end{verbatim}
\end{tcolorbox}

\newpage

\end{document}